\documentclass[5p,times]{elsarticle}

\usepackage{lineno,hyperref}
\usepackage{amssymb}
\usepackage{subcaption}
\usepackage{array}
\usepackage{xcolor}
\usepackage{textcomp}
\usepackage{makecell}
\usepackage{listings}

\usepackage{textcomp }
\usepackage{multirow}
\usepackage{hyperref}
\usepackage[normalem]{ulem}
\usepackage{amsmath}
\usepackage{xurl}
\usepackage{algorithm}
\usepackage[noend]{algpseudocode}
\usepackage{graphicx}
\usepackage{tabularx}
\usepackage{manfnt}
\useunder{\uline}{\ul}{}

\modulolinenumbers[5]

\journal{Future Generation Computer Systems}

\begin{document} \sloppy

\begin{frontmatter}

\title{Automated Evolutionary Approach for the Design of Composite Machine Learning Pipelines}

\author{Nikolay O. Nikitin}
\ead{nnikitin@itmo.ru}

\author{Pavel Vychuzhanin}
\ead{pavel.vychuzhanin@itmo.ru}

\author{Mikhail Sarafanov}
\ead{mik_sar@itmo.ru}

\author{Iana S. Polonskaia}
\ead{ispolonskaia@itmo.ru}

\author{Ilia Revin}
\author{Irina V. Barabanova}
\author{Gleb Maximov}
\author{Anna V. Kalyuzhnaya}
\author{Alexander Boukhanovsky}

\address{ITMO University, Saint-Petersburg, Russia}

\begin{abstract}
The effectiveness of the machine learning methods for real-world tasks depends on the proper structure of the modeling pipeline. The proposed approach is aimed to automate the design of composite machine learning pipelines, which is equivalent to computation workflows that consist of models and data operations. The approach combines key ideas of both automated machine learning and workflow management systems. It designs the pipelines with a customizable graph-based structure, analyzes the obtained results, and reproduces them. The evolutionary approach is used for the flexible identification of pipeline structure. The additional algorithms for sensitivity analysis, atomization, and hyperparameter tuning are implemented to improve the effectiveness of the approach. Also, the software implementation on this approach is presented as an open-source framework. The set of experiments is conducted for the different datasets and tasks (classification, regression, time series forecasting). The obtained results confirm the correctness and effectiveness of the proposed approach in the comparison with the state-of-the-art competitors and baseline solutions.
\end{abstract}

\begin{keyword}
AutoML\sep workflow\sep composite pipeline \sep machine learning \sep evolutionary algorithms \sep WMS
\end{keyword}

\end{frontmatter}


\section{Introduction}


Nowadays, the methods of machine learning (ML) are quite promising approach in various scientific and applied problems. To solve the complicated modeling task with ML models (especially in the cloud and distributed environments), the structure of the model is decomposed to a large number of interconnected blocks that are combined into workflow \cite{visheratin2016workflow}. The common concept of workflow is usually considered as an organization of resources into processes.

\begin{figure*}[t]
\centerline{\includegraphics[width=16cm]{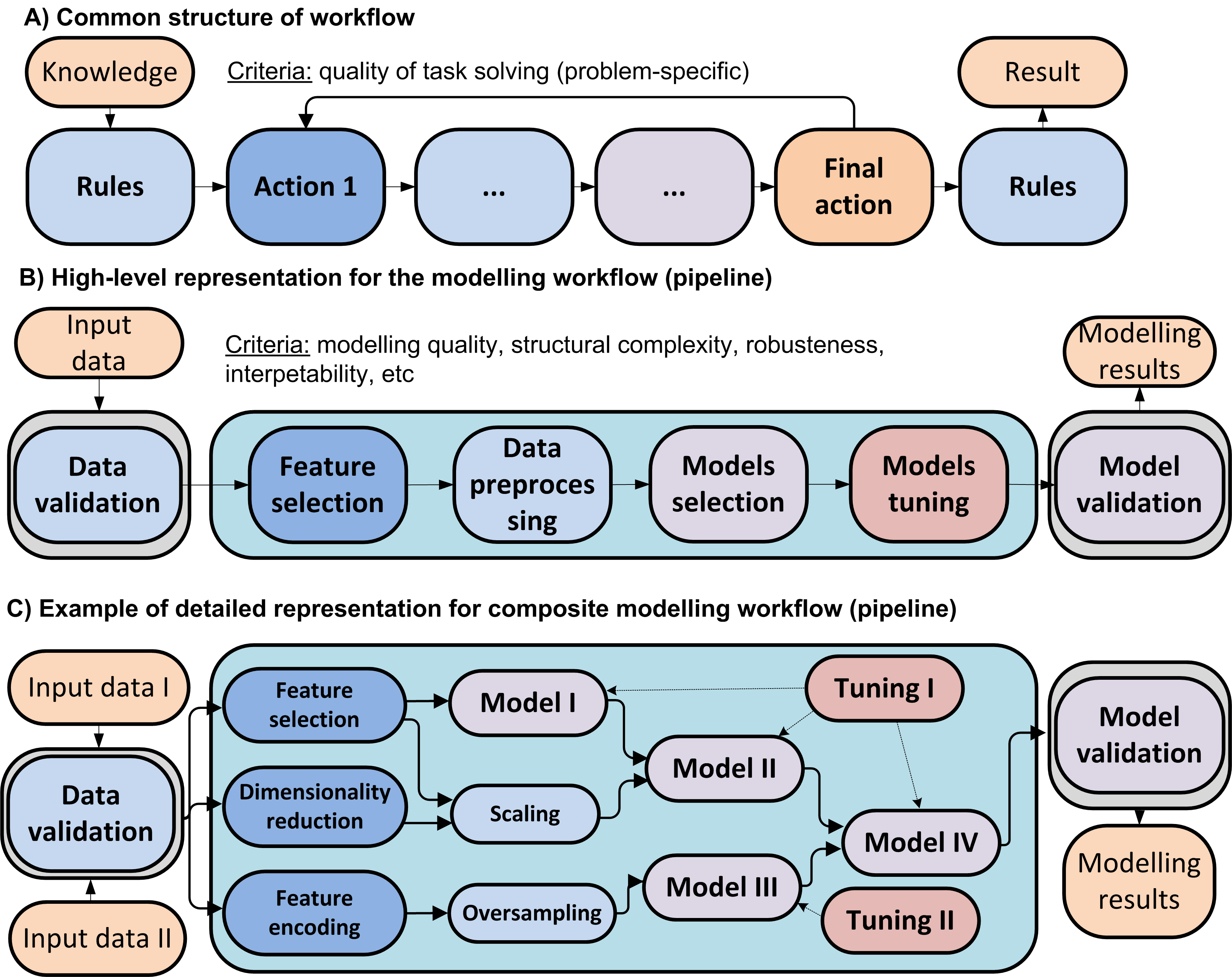}}
\caption{The comparison of different representation workflows: a) the problem-independent representation of the common workflow b) the high-level structure of the machine learning modelling workflow (pipeline). c) the detailed structure of the composite pipeline for machine learning-based modelling. The criteria for the automated design of workflows are highlighted.}
\label{fig_wms}
\end{figure*}

A lot of specialized workflow management systems (WMS) exist to deal with multiple interconnections in sophisticated workflows (an example is presented in Fig.~\ref{fig_wms}a). The main aim of the WMS is to automate the generation or adaptation of the workflow to a specific task. There are a lot of research works devoted to this problem exists \cite{reijers2016effectiveness}, as well as software implementations of WMS \cite{liu2014survey}. However, the problem of workflow identification is far from the final solution \cite{cichocki2012workflow}.

The task of workflow design automation can be divided into two conceptual aspects: the object for optimization and the method for it. The optimization object for the workflow is usually represented as the structure of the graph or nodes in the static graph. The possible methods are workflow scheduling \cite{chirkin2017execution, thekkepuryil2021effective} for execution time minimization or knowledge engineering (for the composite applications with an understandable internal logic) \cite{nasonov2018multi}.
 

It is a complicated task to formalize the objective function for workflow optimization (for example, expert-based identification of complex knowledge graphs can be required \cite{smirnov2015ontological}). It makes it impossible to automate the design for any custom workflow, but the specialized solutions for specific types of workflows still can be proposed \cite{smirnov2014linked}. The workflows that are specialized for modeling tasks are usually referred to as pipelines (linear, variable-shaped \cite{zoller2019benchmark} or composite models \cite{kovalchuk2018conceptual}). In large distributed systems, the modeling pipelines are usually used as part of the composite application that controls the evaluation of the models.

A promising direction of research is the design of machine learning (ML) pipelines, which can be considered as a subclass of computational workflows. The main stages of the typical ML modeling pipeline are represented in Fig.~\ref{fig_wms}b. This pipeline has a linear structure, that makes the automation task relatively simple. It can be solved using basic optimization methods, e.g. grid search \cite{zoller2019benchmark}. The difference from other types of workflows is that the objective for the ML pipelines can be effectively formalized as an estimation of modeling error \cite{xin2021whither}.


In practice, the linear pipelines are mostly applicable to simple benchmarks \cite{AutoML2021}. In practice, it can be hard to build an effective model with raw training data using the pipeline with a static structure. There are a lot of different specialized blocks that can be involved in the pipelines in a parallel way. As an example, data preprocessing, feature selection, dimensionality reduction, gap filling, ensembling, post-processing blocks can be mentioned. An example of the structure of a graph-based (composite) pipeline is presented in Fig.~\ref{fig_wms}c.


It can be seen that there are a lot of complementary blocks involved in the pipeline: e.g. different feature selection algorithms are applied to the same dataset to extract more useful features from the data. However, building composite pipelines (optimal selection and tuning of the blocks and connection between them) is a complicated and time-consuming task even for experts in data science. It is further complicated by the fact that the existing implementations of the WMS for machine learning have a lack of 'intelligent' automation and still require a lot of expert involvement. As the most advanced example of industrial WMS for ML pipelines, the widely-used Azure ML \cite{team2016azureml} can be noted. At the same time, the design of a composite pipeline is a complex problem that involves experts in both data science and the target domain. It raises a lot of issues that make it too complicated to deliver the ML into real-world business processes.

The automating of for ML pipelines design is an especially promising task due to a set of factors:
\begin{itemize}[]
    \item[(a)] the ill-structured domain-specific knowledge usually does not have a critical influence in ML workflows;
    \item[(b)] the objective functions for the workflow optimization can be defined in an implicit and computable way;
    \item[(c)] there are a lot of ML algorithms implemented as open-source libraries, that allow selecting workflow block implementations from a large set of solutions.
\end{itemize}

There are a lot of Automated Machine learning (AutoML) approaches to the automating of ML-based modeling \cite{AutoML2021}. The most common targets of automation are combined algorithm selection and hyperparameter optimization (the CASH acronym is frequently used for it). Besides that, there are several approaches that also take the automation of the feature selection and data preprocessing into account. 


The design of ensembles of the ML models can also be automated using state-of-the-art frameworks. Also, there are a lot of tools that allow solving the Neural Architecture Search (NAS) problem and identifying the most suitable structure of a neural network for a specific problem \cite{elsken2019neural}. However, most of the existing AutoML solutions are quite distanced from real business and industrial applications. Although there are several enterprise-related frameworks that exist, they do not always achieve the appropriate quality of the pipelines. The reason is that AutoML covers (a) and (b) points from the described list of potential effectiveness factors, but point (c) - the flexible selection and design of building blocks is still an unsolved problem here. On contrary (c) is an essential component of many WMS \cite{bolt2016scientific}. 

That is why we decided to develop a flexible approach that combines the main features of AutoML and WMS. It allows us to increase the quality of the modeling of different real-world processes using machine learning and hybrid models with a complex, heterogeneous structure. The second aspect is the open-source implementation of this approach, which makes it possible to use the proposed approach for different purposes in a transparent and reproducible way. For these reasons, the open-ended evolutionary automated modeling techniques and their software implementation in a FEDOT framework are described. Different aspects of composite modeling are discussed: automation of pipelines design, analysis of the obtained pipelines, reproducibility of the pipelines.

The paper is organized as follows: Section~\ref{sec_related} contains the analysis of the existing AutoML solution; Section~\ref{sec_problem} describes the problem statement for the composite machine learning models' design; Section~\ref{sec_approach} provides an extensive description of the proposed evolutionary approach to models' design; Section~\ref{sec_software} describes the open-source software implementation of the proposed approach as a part of the framework; Section~\ref{sec_exp} provides the results of the experiments for different benchmarks; Section~\ref{sec_disc} contains the discussions on the perspectives of modern AutoML and the main conclusions for the paper.

\section{Related works}
\label{sec_related}

\subsection{Design of data-driven modeling pipelines}

There are a lot of mathematical methods for modeling, predicting, and forecasting of various processes that allow building conceptual models on this data \cite{caldwell2013mathematical}. These models can have a different nature. They can be based on physical laws represented as equations that are known a priori, that allow reproducing the underlying nature of the target phenomena. If the specific physical law is unknown, equation discovery methods can be used \cite{hvatov2020}.

In many real cases (e.g. modeling of technical or social systems) there is no physics-based assumption for the model design. In this case, statistical and machine learning (ML) methods have demonstrated their high effectiveness \cite{jordan2015machine} and now they are applicable in various fields \cite{das2017survey}. However, it is not enough just to select a machine learning method to build an effective model - the entire modeling pipeline is required to process the data and models. For example, the quality of the obtained model largely depends on the choice of the data preprocessing approach \cite{zelaya2019towards} or the effectiveness of hyperparameters tuning \cite{probst2019tunability}.

There is a lot of data processing, model selection, or hyperparameters tuning methods exist. However, due to the inhomogeneity, multi-modality, and stochasticity of the raw un-processed datasets, a promising way to improve the quality of the machine learning models is to build ensembles of models \cite{sagi2018ensemble} and combine imprecise predictions of several ML models (or the instances of same ML model with different data as inputs) into one, better prediction. Different approaches to ensembling exist:  stacking \cite{pavlyshenko2018using}, bagging and boosting \cite{gonzalez2020practical}, hybridization \cite{ardabili2019advances}.

Bagging involves dividing the original training set into many sub-samples, each of which is used for teaching a separate model in an ensemble. As an example, the random forest model can be used: a lot of independent decision trees are trained on bagging, and in the final forecast their results are combined. Boosting is based on sequential training of models, each of which is aimed at reducing the error of the predecessor. By default, this ensemble method does not impose restrictions on the types of models used. As an example, boosting ensembles based on decision trees has gained the greatest popularity in machine learning. The widely-used open-source solutions for boosting are XGBoost, LightGBM, and CatBoost. For stacking, the final ensemble can have a more complex structure, which can be represented as a directed graph, where nodes are ML models, and connections are data streams \cite{konstantinov2020generalized}. For example, if the default bagging-based ensemble uses the voting classifier \cite{ruta2005classifier} to combine model forecasts, the voting logic can be implemented in a separate ML model.

Ensemble-based modeling methods are also used in the fields of environmental process analysis. The numerical models based on the equations of mathematical physics can be included in ensembles to increase the simulation quality. Models in the ensemble can vary in training data and the internal configuration: meta-parameters of numerical models, boundary and initial conditions \cite{vychuzhanin2019robust}.

The hybrid modeling is based on the ensembling of the data-driven and numerical models \cite{sun2020comprehensive}. In such a way, the results of mathematical physics models can be improved using data-driven model predictions in various subject domains \cite{zhang2019hybrid}. The main disadvantage that limits the use of this technology in composite modeling is the lack of a universal interface for domain-specific models, which significantly complicates their use in composite pipelines.

The ensemble-based methods have a lot of implementations. For example, the described ensemble methods are implemented within the machine learning frameworks, e.g. Scikit-Learn \cite{pedregosa2011scikit}. Nevertheless, the problem of their combination as a part of a modeling pipeline still has no satisfactory solution. In this paper, we used the concept of a composite model (described in \cite{kalyuzhnaya2020automatic}) to represent the various structures of pipelines in a unified form. Data-driven models can not be distinguished from the corresponding pipeline, so the composite model and composite pipeline can be considered synonymous in this case.

From the perspective of applied data science, the modeling approach as a numerical method is inseparable from its software implementation. That is why we devoted the second part of this section to the review of state-of-the-art frameworks and tools for data-driven pipelines design and evaluation.

\subsection{Automated tools for machine learning pipeline design}

In the field of modern ML, there are a number of various tools, frameworks, and platforms that allow building data-driven models of the different processes and events of the real or virtual world. In the frame of this paper, we focused on the analysis of existing solutions of AutoML-related products from both scientific and enterprise domains. The aim of the analysis is to identify the main advantages and disadvantages of existing tools of automated modeling in order to substantiate the relevance of the proposed approach to the model design.

Although many variations of the AutoML tools exist, many of them are based on the same ML frameworks that are used for the ML part of the modeling itself. Each AutoML tool can be described in the following way: class of problems that can be solved (classification, regression, etc); types of data that can be processed (tables, images, texts, etc); parts of the ML pipeline that can be automated; types of models that can be used in the automated design; optimization approach used to automate the modeling; availability of the open-source implementation. The results of this comparison for a set of state-of-the-art solutions for pipeline automation are presented in Table~\ref{tab_automl_surv}.

Here, several issues of AutoML tools could be noted. Firstly, most of the solutions support only fixed-shaped pipelines: its structure is either predefined, or the suitable structure is selected from a set of the candidates. The experiments based on well-known benchmark datasets show that simple linear pipelines generated by AutoML can be effective enough \cite{erickson2020autogluon}. However, it is not clear how the resulted solutions are robust for the variability in the input data.

The second concern about the modern AutoML is the single modality of used datasets. In most real-world tasks, data scientists deal with heterogeneous data, including texts, numerical and categorical features, as well as time-series. In most cases, the multi-modal pipeline design is identified manually, and most of the AutoML solutions support automation only for single modality pipelines. Even if the approach of multi-modal data-driven modeling is proposed \cite{yin2020identifying}, it is usually implemented via a single model, such as an auto-encoder neural network. At the same time, the described concept of multi-modality suits well into the composite pipelines.          

\begin{table*}
\caption{The comparison of the main aspects of several state-of-the-art open-source AutoML tools. Short references to the repositories in \url{github.com} are provided.}
\centering
\refstepcounter{table}
\label{tab_automl_surv}
\begin{tabular}{|c|c|c|c|c|c|c|} 
\hline
\textbf{Tool}                                        & \textbf{ Pipeline } & \begin{tabular}[c]{@{}c@{}}\textbf{ Optimisation}\\\textbf{algorithm }\end{tabular} & \textbf{ Input data }                                           & \textbf{ Scaling }                                                & \begin{tabular}[c]{@{}c@{}}\textbf{ Additional }\\\textbf{features }\end{tabular} & \textbf{ GitHub repository}                               \\ 
\hline
TPOT                                                  & Variable            & GP                                                                                  & Tabular                                                         & \begin{tabular}[c]{@{}c@{}}Multiprocessing,\\~Rapids\end{tabular} & \begin{tabular}[c]{@{}c@{}}Code \\generation\end{tabular}                         & \url{EpistasisLab/tpot}            \\ 
\hline
H2O                                                   & Fixed               & Grid Search                                                                         & Tabular, Texts                                                  & Hybrid                                                            & -                                                                                 & \url{h2oai/h2o-3}                  \\ 
\hline
\begin{tabular}[c]{@{}c@{}}Auto\\Sklearn\end{tabular} & Fixed               & SMAC                                                                                & Tabular                                                         & -                                                                 & -                                                                                 & \url{automl/auto-sklearn}          \\ 
\hline
ATM                                                   & Fixed               & BTB                                                                                 & Tabular                                                         & Hybrid                                                            & -                                                                                 & \url{HDI-Project/ATM}              \\ 
\hline
\begin{tabular}[c]{@{}c@{}}Auto\\Gluon\end{tabular}   & Fixed               & Fixed                                                                               & \begin{tabular}[c]{@{}c@{}}Tabular, Images,\\Texts\end{tabular} & -                                                                 & \begin{tabular}[c]{@{}c@{}}NAS, \\AWS integration\end{tabular}                    & \url{awslabs/autogluon}            \\ 
\hline
LAMA                                                  & Fixed               & Fixed, Optuna                                                                       & Tabular, Texts                                                         & -                                                                 & Profiling                                                                         & \begin{tabular}[c]{@{}c@{}}\url{sberbank-ai/}, \\\url{LightAutoML}\end{tabular}  \\ 
\hline
NNI                                                   & Fixed               & Bayes                                                                               & Tabular, Images                                                 & \begin{tabular}[c]{@{}c@{}}Hybrid, \\Kubernetes\end{tabular}      & NAS, WebUI                                                                        & \url{microsoft/nni}                \\
\hline
\end{tabular}
\end{table*}

Also, there are a lot of more specific frameworks that are oriented towards using one single modeling task: the EPDE framework for the automated design of models based on differential equations \cite{maslyaev2020data}, the ClusterEnsembles framework for the automation of clustering \cite{strehl2002cluster}, AutoTS for time series forecasting \cite{khider2019autots}, etc.

It can be seen that a large number of well-developed frameworks exist. Nevertheless, there is no ready-to-use solution that can be applied as a universal tool to automate the generation of a data-driven model with a complex structure.

\subsection{Workflow management systems}

There is a lot of common-purpose and specialized WMS exists \cite{yu2005taxonomy}. The main features of WMS are: (a) scheduling of the workflow execution \cite{visheratin2016workflow}; (b) processing of different task-specific building blocks \cite{bolt2016scientific}; (c) involvement of the domain knowledge into the workflow execution and design \cite{smirnov2015ontological}. These features can be useful not only for classical workflows but also for machine learning pipelines. However, the effectiveness of the WMS for machine learning is limited due to the lack of automation of pipeline design.

To partially solve the problems described in the section, we proposed an approach that allows solving the data-driven design problem in an automated, flexible, and interpretable way using multi-objective evolutionary optimization and features derived from both AutoML and WMS. To make the approach available for the broad community and adaptable to new tasks, we implemented it as a part of the open-source framework described in Section~\ref{sec_software}.

\section{Problem statement}
\label{sec_problem}

Several different classes of modeling pipelines can be distinguished. In \cite{zoller2019benchmark}, the fixed structure pipelines and variable shaped pipelines are discussed. However, the continuous development of the data-driven modeling approach raises a set of issues, that can be solved using pipelines with an even more complex and heterogeneous structure.

Modeling pipelines can be represented as graphs with different properties. The simplest linear pipelines can be described as a path graph, ensemble-based pipelines can be described as a minimum spanning graph, and composite pipelines are directed acyclic graphs with a single limitation - the existence of the final node (the destination for all data flows). This classification is detailed in Table~\ref{tab_graphs}.

\begin{table}[h!]
  \renewcommand\tabularxcolumn[1]{m{#1}}
  \newcolumntype{Y}{>{\centering\arraybackslash}X}
  \centering
  \begin{tabularx}{\columnwidth}{|c|Y|Y|}
    \hline
    Structure & Graph type & Represented modelling workflow \\ \hline
    \begin{minipage}{.15\textwidth}
      \includegraphics[width=\linewidth]{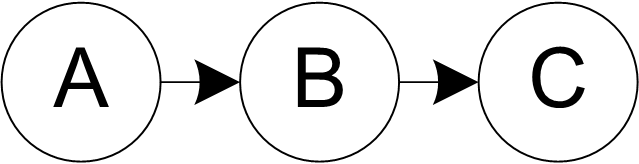}
    \end{minipage}
    &
    Path graph (only one parent for each node)
    & 
    Linear pipelines
    \\ \hline
    \begin{minipage}{.15\textwidth}
      \includegraphics[width=\linewidth]{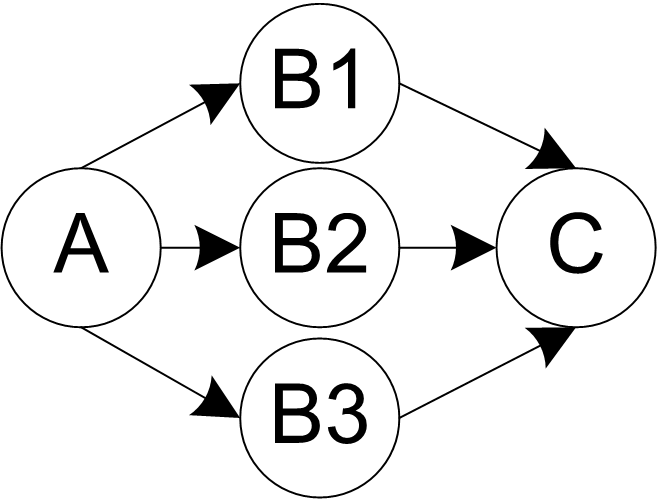}
    \end{minipage}
    &
    Minimum spanning graph (all paths between all pairs of nodes have the same length)
    & 
    Ensemble pipelines (low variability)
    \\ \hline
    \begin{minipage}{.15\textwidth}
      \includegraphics[width=\linewidth]{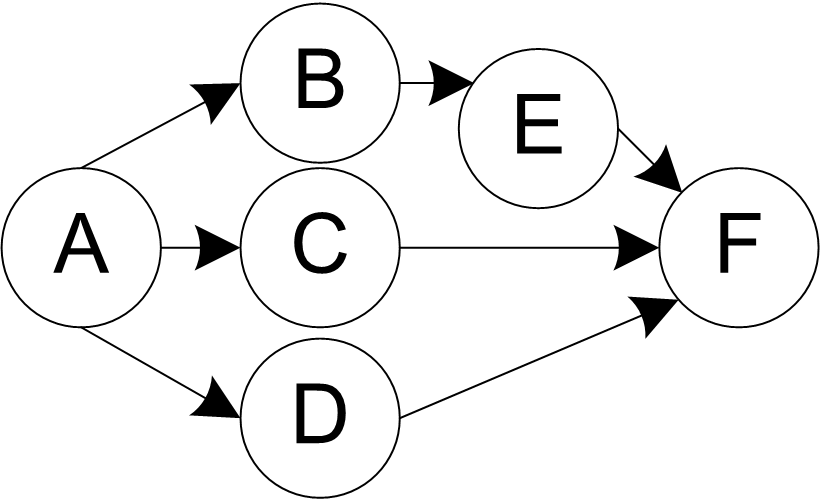}
    \end{minipage}
    &
    Directed acyclic graph (with all paths finishing in the final node)
    & 
    Composite pipelines (high variability)
    \\ \hline
  \end{tabularx}
  \caption{The description of different graph representations for modelling workflows}
  \label{tab_graphs}
\end{table}

In this paper, we consider composite pipelines (also referred to as composite models \cite{kalyuzhnaya2020automatic}) as a more perspective approach. A composite pipeline is a heterogeneous data-driven model with a graph-based internal structure, within which several atomic functional blocks (models and data-related operators) can be identified. The main concepts of well-known variable-shaped pipelines (e.g. implemented in \cite{olson2016tpot}) are close to composite pipelines, but several differences should be noted. 

Firstly, composite pipelines can contain more than one model aimed for different purposes (classification, clustering, forecasting, etc.) and use data with different types (text, images, tables, etc.). For instance, some part of the composite pipeline is aimed to forecast time-series based on given prehistory. And the output is used as an additional feature for a final classification task. Thus, in composite pipelines, several data sources (even with different types) can be involved. 

Secondly, all predictive models and data processing algorithms are represented as functional blocks ${A}_{i}$ and processed in a unified form as a part of the pipeline. Therefore the proposed approach operates with various types of atomic blocks. Also, for composite pipelines it is possible to introduce a fractal property: a composite model can be used as an atomic block thus it can be included in another composite model.         

From the mathematical point of view, the structure of composite pipeline $P$ can be described as a directed acyclic graph (DAG) $G$. In this case, graph representation ${P}^{G}$ can be formalized as follows:

\begin{equation}
\begin{aligned}[t]
	{{P}^{G}}=\left\langle {{V}_{i}},{{E}_{j}} \right\rangle =\left\langle {{A}_{i}},{{\{{{H}_{{{A}_{i}}}}\}}_{k}},{{E}_{j}} \right\rangle,
\end{aligned}
\end{equation}	
where $V$ are the vertices with a complex structure that can be represented as a tuple $\left\langle {{A}_{i}},{{\{{{H}_{{{A}_{i}}}}\}}_{k}}\right\rangle$ with modelling or preprocessing functional blocks ${A}_{i}$ and their hyper-parameters ${{\{{{H}_{{{A}_{i}}}}\}}_{k}}$. Directed edges $E$ represent the data flow between functional blocks.

In this case, the optimization task for the pipeline structure can be described as follows:

\begin{equation}
	{f}^{max}\left( {{P}^{*}} \right)=\underset{P \in \mathbb{P}}{\mathop{\max }} \,{f}\left( P|{{T}_{gen}}\le {{\tau }_{g}} \right),
	\end{equation}
where $f$ is the objective function that characterizes the modelling quality for the considered workflow and ${f}^{max}$ is the maximal value of the fitness function obtained during optimization, $\mathbb{P}$ is a set of possible pipeline structures (search space), ${{T}_{gen}}$ is time spent for workflow design, ${\tau}_{g}$ - is time limit.

The application of the proposed approach to real-world problems is quite a promising research direction. At the same time, the described integration of the WMS and AutoML features raises several issues that should be solved during the implementation of the approach. In the paper, we try to propose possible solutions to them and analyze their effectiveness.

\subsection*{Issue 1: Do composite pipelines have an advantage against more simple pipelines in real tasks?}
\label{iss_composite}

There are a lot of examples of the modeling workflows application to different real-world problems \cite{atkinson2017scientific}. Most of them are designed by domain experts or data scientists (using pipeline automation solutions that are close to WMS). At the same time, the most of AutoML tools allow building simple pipelines with near-fixed structure. The practical applicability of the composite pipelines to the variety of modeling tasks (classification, regression, time series forecasting, clustering) is still unclear since the impact of time and resources limitations can overcome the potential effort from the more flexible structure. The corresponding experimental results are not systematized in the literature. In order to substantiate the proposed approach, it is necessary to provide the experimental evaluation of the composite pipelines against baselines and alternative methods of pipeline design.

\subsection*{Issue 2: Which blocks can be used as a part of a composite pipeline?}
\label{iss_blocks}

The graph-based representation of the composite pipeline allows representing a wide range of possible configurations of data-driven models \cite{qi2020graph}. However, the possible types of functional blocks are quite limited in the existing AutoML approaches. The WMS allows integrating different blocks into workflows, but there is still no unified way to represent different models and operations that can be used during the automated design of the composite pipelines approach described in the paper.

\subsection*{Issue 3: How to combine the AutoML and WMS approaches for the automated pipeline design?}
\label{iss_design}

 There is no oblivious way to combine the best features of approaches implemented in existing AutoML and WMS. The tools for automated machine learning (AutoML) are mostly focused on linear or sub-linear pipelines (see Sec.~\ref{sec_related}), so the advanced approach (i.e. genetic algorithms) can be redundant \cite{zoller2021incremental}. Most WMS are not aimed to solve the design problems in an automated way and focused on decision-making support for the expert with partial automation only \cite{knyazkov2012clavire}. So, there is no ready-to-use concept for an implementation of a hybrid automated modeling approach that can be efficiently applied to composite pipelines. To build the modeling pipelines in an effective, robust, and controllable way, a suitable automatic method should be chosen or developed. Also, it should be applicable for all types of pipelines described in Table~\ref{tab_graphs} to avoid both over-complicated and over-simplified solutions.

\subsection*{Issue 4: How to analyze the structure of pipeline?}
\label{iss_interp}

The WMS-like approach to the pipeline design raises the structural analysis problem for the pipeline. The structure of the modeling pipeline consists of many different blocks and connections between them. If the composite pipeline is obtained using automated design methods (see \hyperref[iss_design]{Issue 3})), its analysis can be confusing for the expert. To simplify it, an automated approach that allows estimating the impact and importance of each structural block is required. The existing methods allow analyzing the hyperparameters of the models in blocks, but there is no such approach for structural analysis. 

\subsection*{Issue 5: How to tune the hyperparameters in composite pipelines?}
\label{iss_tuing}

Fine-tuning of the hyperparameters in the composite pipeline usually differents from that are using in existing AutoML tools for single machine learning models \cite{kalyuzhnaya2021towards}. As an example, the data preprocessing block also require appropriate tuning, but the quality metric evaluation can not be provided without connection with the models. To improve the efficiency of the approach described in the paper, a unified strategy of hyper-parameter tuning that can be effective for different modeling tasks and types of data is required.

\subsection*{Issue 6: How to reproduce the results of composite modeling?}
\label{iss_repr}

The reproducibility of machine learning pipelines and related experiments is a critical problem in many scientific and industrial applications \cite{sugimura2018building}. The complex structure of the composite pipelines raises the additional aspects of reproducibility and reputability problems. There are different ways or models export exist in the existing AutoML tools: code generation, binary files, etc. Some WMS can define the domain-specific language (DSL) for the description of the workflow \cite{cordasco2020toward}. However, the application of full-scale custom DSL can be redundant for the modeling pipelines. So, the specialized method that allows exporting and importing the composite pipelines in a simple, human-understandable, and reliable way should be implemented.

To sum up, the problem statement for the proposed approach to the composite data-driven modeling pipelines design consists of \hyperref[iss_design]{Issue 1}) - \hyperref[iss_design]{Issue 6}). The paper is devoted to the justification of the proposed solutions to these issues, as well as the description, analysis, and validation of the implemented approach.

\section{Design of the composite data-driven workflows}
\label{sec_approach}

This section is devoted to the various aspects of the proposed approach, that allows designing the composite data-driven workflows in an automated way. Since it is based on a concept of a combination of best practices from both AutoML and WMS, both model-related and automation-related aspects of the pipeline design problem were considered as a subject for analysis, validation, and justification.

\subsection{Pipelines, models, and operations}

As it was noted in Sec.\ref{sec_problem}, the pipeline can be represented as an acyclic directed graph. The basic abstractions of the composite pipeline are:
\begin{itemize}
  \item \textbf{Operation} - an operation is an action that is performed on the data: an algorithm for data transformation or machine learning model that produces predictions;
  \item \textbf{Node} is a container in which an operation is placed. There can only be one operation in one node. A \textbf{Primary} node accepts only raw data, and a \textbf{Secondary} node uses the predictions of the previous level nodes as predictors;
  \item \textbf{Pipeline} is the implementation of the modeling pipeline (with defined fit/predict operations) represented as a graph that consists of nodes (as vertices) and data flows (as edges).
\end{itemize}

To separate pipeline search space, we divide (see Table~\ref{tab_operations}) the building blocks for pipelines into two major groups: a) models and data operations, that are typical blocks for AutoML techniques; b) task-specific models (specific statistical models, equation-based-models, etc) and operators for data flow (merging, target decomposition, data source, stochastic data generators, etc). The blocks for modeling contain operations that transform features into predictions. The data preprocessing operations can change features but do not approximate the relationship between features and predictions. The data flow operations can transform (or generate) both features and targets.

\begin{table}
\centering
\caption{Types in the building blocks can be used in the proposed structure of the modeling workflow: a) blocks that usually using in the pipelines generated by AutoML; b) blocks usually using in the manually-designed modeling workflows}
\label{tab_operations}
\begin{tabular}{|c|c|c|c|} 
\hline
\begin{tabular}[c]{@{}c@{}}Building \\blocks \\types\end{tabular}    & Description & Example & Metadata \\ 
\hline
\multicolumn{4}{|c|}{a) AutoML-specific types}\\ 
\hline
\begin{tabular}[c]{@{}c@{}}ML \\models\end{tabular}                  & \begin{tabular}[c]{@{}c@{}}Transform \\features into \\predictions\end{tabular} & \begin{tabular}[c]{@{}c@{}}Ridge\\regression\end{tabular}                     & \begin{tabular}[c]{@{}c@{}}\#simple,\\\#linear, \\\#interpret-\\able\end{tabular}     \\ 
\hline
\begin{tabular}[c]{@{}c@{}}Data\\processing\\operations\end{tabular} & \begin{tabular}[c]{@{}c@{}}Modify \\data without \\prediction\end{tabular}                                          & \begin{tabular}[c]{@{}c@{}}Outlier\\filtering\end{tabular}                    & \#non-linear                                                                          \\ 
\hline
\multicolumn{4}{|c|}{b) Workflow-specific types} \\ 
\hline
\begin{tabular}[c]{@{}c@{}}Task-specific\\models\end{tabular}        & \begin{tabular}[c]{@{}c@{}}Suitable for \\specific\\data types\\and tasks\end{tabular}                                           & \begin{tabular}[c]{@{}c@{}}Equation-based\\runoff \\model\end{tabular}        & \begin{tabular}[c]{@{}c@{}}\#interpret-\\able,\\\#time-series\\specific\end{tabular}  \\ 
\hline
\begin{tabular}[c]{@{}c@{}}Data flow \\operation\end{tabular}        & \begin{tabular}[c]{@{}c@{}}Modify \\target based \\on previous\\model \\predictions \end{tabular} & \begin{tabular}[c]{@{}c@{}}Merge data \\from different \\sources\end{tabular} & \#non-default                                                                         \\
\hline
\end{tabular}
\end{table}

The description of the operations is stored in JSON files. Tags are used to control the "task to solve - operation to use" match. An example of several operations with the tags associated with it can be seen in Table~\ref{tab_operations}.

The tag filtering mechanism allows flexible selecting of the appropriate blocks for composing pipelines. Thus, it is possible to use any combination of the proposed operations. For example, it is possible to create pipelines only with ML models, or only with linear models and simple preprocessing methods.

In the proposed approach a uniform data flow strategy should be used regardless of the specific blocks in the pipeline. There are several ways to represent data flow in the pipeline. The first one is the blocks that use the parent models' predictions as input. The main disadvantage is the restriction of the obtained decision boundary variability since the number of input features is quite small. The second implementation always passes the data to the input of any model (for example, this technique is implemented in the TPOT tool as a stacked estimation block), but it also has a lack of flexibility.

\begin{figure*}[h]
\centerline{\includegraphics[width=1.0\textwidth]{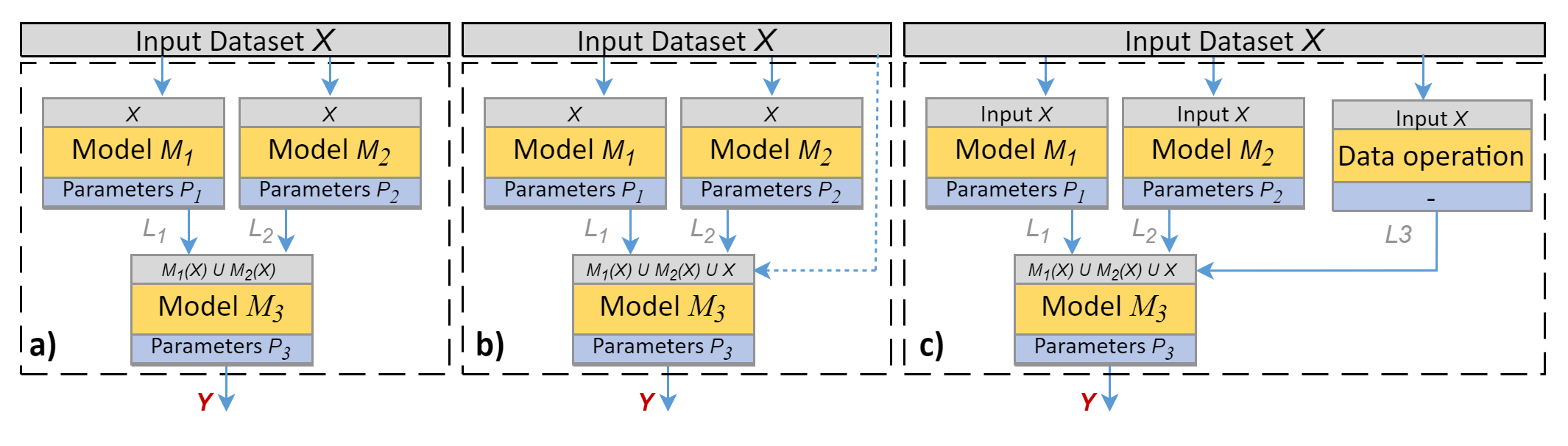}}
\caption{The comparison of different strategies of data flow management for the secondary nodes: a) sequential ; b) direct; c) adaptive.}
\label{fig_enrich}
\end{figure*}

For the genetic programming paradigm, we can implement input data as an independent pseudo-block. In this case, the input of any model can be enriched by the data (if it is useful for the increasing of the fitness metric value). The comparison of several strategies for the management of data flows in the pipeline is provided in Fig.~\ref{fig_enrich}.

Data operations can be considered as data-specific blocks that can potentially increase the accuracy of the final forecast. For time series, it can be moving average filtering, for tables - outliers detection based on the random sample consensus algorithm, or a recursive feature elimination (RFE) for feature selection. Also, normalization, dimensionality reduction, encoding, imputation, and other operations with data can be included in the pipeline design task as building blocks. For this reason, the strategy that allows processing various building blocks and connections between them (similar to WMS) is considered the most flexible and used as a part of the proposed approach for pipeline design.

\subsection{Automated evolutionary design}
\label{subsec_evo}

\begin{figure*}[t!]
\centerline{\includegraphics[width=1.0\textwidth]{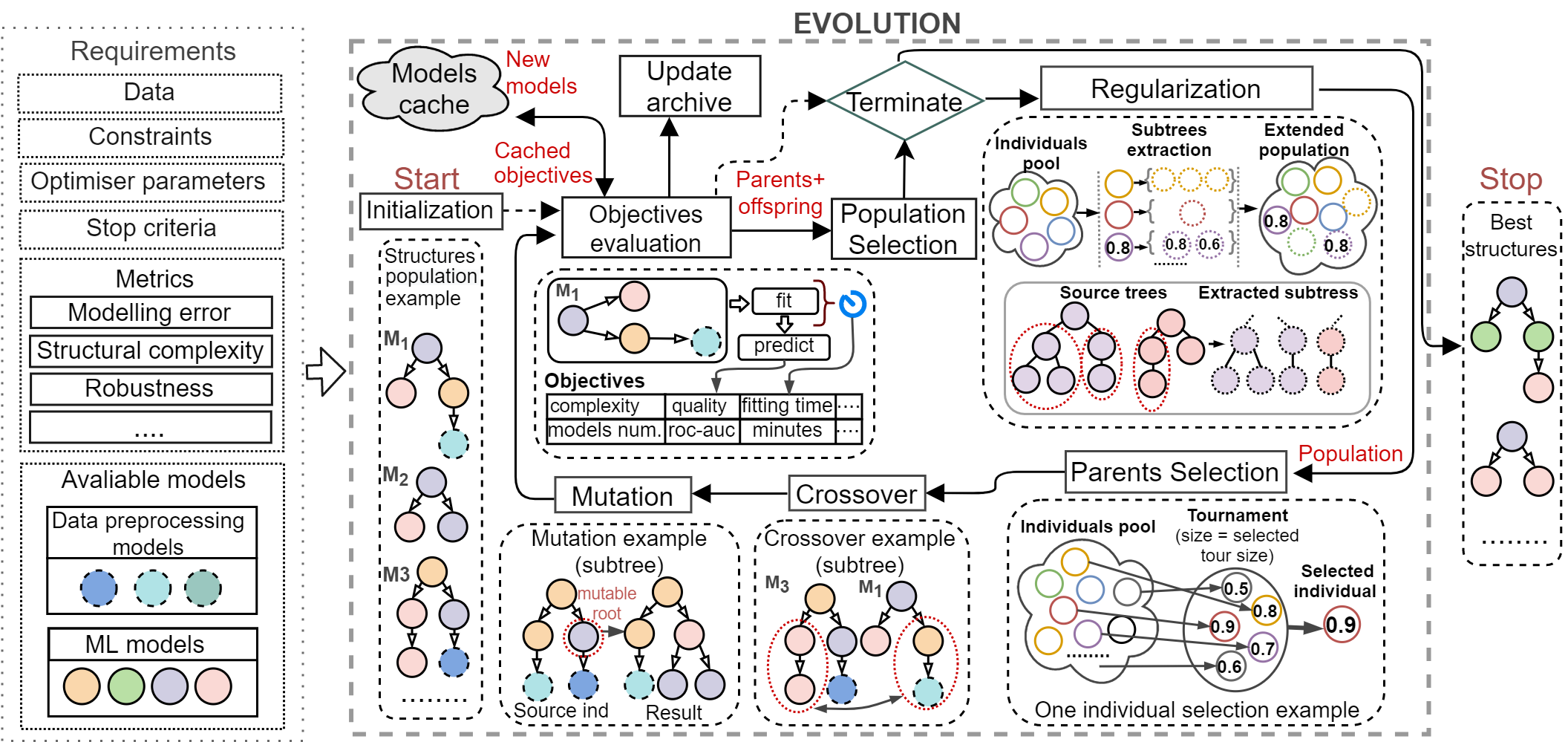}}
\caption{The concept of the multi-objective evolutionary design of the composite machine learning workflows (pipelines) implemented as a part of proposed approach.}
\label{fig_evo}
\end{figure*}

The procedure of the automated design allows selection of model structure for certain input data automatically and consists of two steps:

\begin{itemize}
  \item \textbf{Composition} - the process of finding a pipeline structure. By default, the framework has an evolutionary algorithm responsible for this, in which optimization is performed using genetic selection, crossover, and mutation operators (see Fig.~\ref{fig_evo}). At this stage, operations in nodes are changed, subtrees are removed, added, or extended for each individual in a population. Each operator has a performing probability. Hyperparameters of operations in nodes are also can be mutated. Mutation operators are generally subdivided into two groups: for exploration and for exploitation aims. The first group grows new parts of pipelines (sometimes replacing old selected parts). The second performs local changes or reducing pipelines' parts.;
  \item \textbf{Hyperparameter tuning} is a process in which the structure of a pipeline does not change, but only the hyperparameters in the nodes are tuned. This step is started after the composition is finished.
\end{itemize}

The high-level pseudocode of the evolutionary design approach is presented in Alg.~\ref{alg_gp}. It is necessary to load the database, define the search space, objectives, termination criteria, and evolutionary search algorithm hyperparameters before the optimization. By default, the first population of pipelines is generated randomly, but it is possible to add existing baselines or previously obtained AutoML solutions as an initial assumption to improve the convergence of the search process. We should make every attempt to reuse the existing knowledge because starting the search from scratch is usually expensive. During the entire evolutionary process, the algorithm measures objectives for each new generated pipeline (if it is not contained in cache). A training data sample is used to obtain the structure and parameters of each new composite model. After that, a test sample is used to evaluate metrics (see Alg.~\ref{alg_gp}, Evaluation procedure). The search process stops if one of the stop criteria is satisfied. 

Adaptive evolutionary schemes (as described in \cite{evans2020adaptive}) assume hyperparameters (such as population size, offspring size, rates of evolutionary operators,constraints, e.t.c.) adaptation during the evolution depending on the diversity of the population and its convergence speed (Alg.~\ref{alg_gp}, line~\ref{alg:hyperparams}). 

Besides the user-defined constraints (including structural-based, e.t.c.) used in evolutionary optimization, each new individual produced by crossover or mutation procedure is checked for the rules violation defined based on object representation type (e.g., DAG can not have isolated nodes). More details in Alg.~\ref{alg_gp}, line~\ref{alg:valid}.

In Fig.~\ref{fig_evo} the scheme of the evolutionary design approach is presented. Each evolutionary procedure is demonstrated in simplified form to improve readability. In practice, few different types of each evolutionary operator can be used in the algorithm. In this case, operator type can be selected randomly with specified probabilities. Operators can have either equal probabilities or be assigned by the special evolutionary scheme. Such scheme performs the dynamic adaptation of operators’ probabilistic rates on the level of the population (\cite{semenkina2014hybrid}). In \cite{nikitin2020structural} examples of evolutionary operators for the graph structures (crossovers, mutations, and regularization) with a detailed description. 

\begin{algorithm*}
\caption{Pseudocode of evolutionary algorithm that is implemented as a part of composition module}

\begin{algorithmic}[1]

\Procedure{EvolutionaryComposer}{}
    \State \underline{Input:} objectiveFunctions = $\left\{qualityObj,  complexityObj, ...\right\}$, operations = $\left\{models, dataOperations, ... \right\}$, stopCriterions =
    \State = $\left\{timeLimit, generLimit, ...\right\}$, evoOperators = $\left\{crossoverTypes, mutationTypes, selectionTypes, regularTypes, ...\right\}$,
    \State constraints = $\left\{structuralConst, optimiserConst, ...\right\}$, operatorsRates, data, popSize, initialPipelines, evaluatedPipelinesCache
    \State \underline{Output:} array with nondominated models (Pareto frontier)
    \State $pop\gets \Call{InitPopulation}{models, popSize, structuralConst, initialPipelines} $
    \State $pop\gets \Call{Evaluation}{pop, objectiveFunctions, data}$
    \While{not \Call{Terminate}{stopCriterions}}
        \State $\triangleright$ \textit{In adaptive evo. schemes hyperparameters (e.g. popSize,rates,constraints, e.t.c.) update during evolution}
        \State \Call{UpdateHyperparams}{evolutionaryScheme} \label{alg:hyperparams} 
        \State $pop \gets \Call{Regularization}{pop, regularTypes}$
        \State $\triangleright$ \textit{Pareto frontier contains only one best individual in the case of a single-objective algorithm}
        \State $pareto, parents \gets \Call{UpdatePareto}{pop}, \Call{ParentsSelection}{pop, offspringSize, selectionTypes}$
        \State $offspring \gets \Call{Reproduction}{parents, crossoverTypes, mutationTypes,operatorsRates, constraints, models}$
        \State $pop \gets \Call{Evaluate}{offspring, objectiveFunctions, data, evaluatedPipelinesCache}$
        \State $pop \gets \Call{Selection}{pop\cup offspring, popSize, selectionTypes}$
    \EndWhile
    \State \textbf{return} $pareto$ 
\EndProcedure
\Procedure{Evaluation}{}
    \State \underline{Input:} pop, objectiveFunction, data, cache
    \For {$ind$ in $pop$}
        \State $mlModel\leftarrow$ \Call{Convertor}{$ind$}
        \State $\triangleright$ \textit{Model fitting and obtaining its prediction}
        \State $prediction \leftarrow$ \Call{FitPredict}{$mlModel$,data, cache} 
        \State $\triangleright$ \textit{Metrics evaluation and recording their values to corresponding fields of individual}
        \State \Call{MetricsEvaluation}{$ind$,$prediction$,$objectiveFunctions$ } 
    \EndFor
\EndProcedure
\Procedure{Reproduction}{}
    \State \underline{Input:} parents, crossoverTypes, mutationTypes, operatorsRates = $\left\{crossoverRate, mutatioRate\right\}$, constraints, models
    \State \underline{Output:} array with new individuals
    \State $offspring \gets \left\{\right\}$
    \For {$parent1, parent2$ in $parents$}
        \State $\triangleright$ \textit{Random choice of evolutionary operators types from avaliable}
        \State $crossover \gets \Call{RandomChoice}{crossoverTypes, crossoverRate}$
        \State $mutation \gets \Call{RandomChoice}{mutationTypes, mutationRate}$
        \State $\triangleright$ \textit{Each new individual is checked for the rules which are defined on the basis of object representation type}
        \While{not $\Call{Validation}{newInds}$} \label{alg:valid}
            \State $newInds \gets \Call{crossover}{parent1, parent2, constraints}$
            \State $newInds \gets \Call{mutation}{newInds, constraints, models}$ 
        \EndWhile
        \State $offspring \gets offspring\cup newInds$
    \EndFor
\State \textbf{return} $offspring$ 
\EndProcedure
\label{evo_params_adapt}
\end{algorithmic}
\label{alg_gp}
\end{algorithm*}

\subsection{Hyperparameters tuning strategies}

During the optimal pipeline structure search, the evolutionary algorithm uses mutation operators. The hyperparameters of the nodes can be configured using specialized operators, but the simultaneous search for the optimal pipeline topology and parameters can cause convergence problems. So, the pipeline obtained after the evolutionary design stage is still can be not optimal due to improperly configured hyperparameters. To achieve the appropriate modeling quality, the hyperparameters of each operation should be fine-tuned. Such tuning can be conducted in different ways.

Since the result of the pipeline, design has a graph-based structure, it is possible to tune the nodes sequentially or simultaneously. At the same time, for tuning, regardless of the strategy, the "black box" Bayesian optimization algorithm is used. So, there are several candidate strategies was proposed:

\begin{itemize}
  \item Serial isolated tuning - approach to tuning hyperparameters in nodes, where the hyperparameters are tuned sequentially for each node in the pipeline separately. In this case, the hyperparameter optimization is performed in isolation for each node. The data is not passed through the pipeline, but only through the corresponding subtree. The error is minimized at the specific node for which the parameters are selected;
 \item Simultaneous tuning - approach in which hyperparameter optimization is performed simultaneously for each node in the pipeline. Thus, the entire composite pipeline is optimized as a black-box. The error is minimized for the entire pipeline;
 \item Sequential tuning - approach is similar to "serial isolated tuning", but the difference is that if the hyperparameters are optimized at one node, the quality metric is recalculated for the entire pipeline. The error is minimized for the entire pipeline.
 \end{itemize}

\begin{figure}[t!]
\centerline{\includegraphics[width=0.45\textwidth]{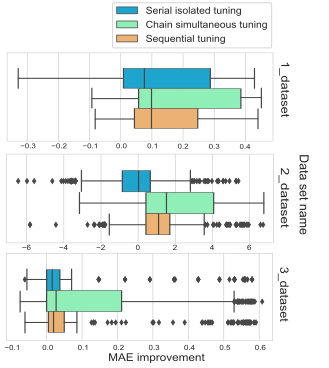}}
\caption{Comparison of the three proposed optimization algorithms in terms of improving the quality metric of the composite pipeline.}
\label{fig_tuning_comprasion}
\end{figure}

We compare these strategies on two types of problems: classification and regression. For each problem type, three composite pipelines with different complexity were analyzed. Each experiment was repeated 100 times to analyze the both quality and stability of the tuning. The number of iterations for tuning is 100. The MAE and ROC-AUC metrics were measured before and after setting the parameters. The results of comparing the algorithms for the regression problem can be seen in Fig.~\ref{fig_tuning_comprasion}.

In the regression problem, simultaneous tuning proved to be the most effective approach, as it provided the greatest increase in accuracy and provided more reliable solutions. For the classification problem, the sequential tuning and simultaneous tuning approaches showed the same result. Serial isolated tuning in both problems proved to be the most inaccurate approach, but the most computationally cheap of the presented ones. For these reasons, we used simultaneous tuning as a part of automated pipeline design during the experimental studies described in Sec.~\ref{sec_exp}.

\subsection{Analysis and improvement of the pipeline structure using sensitivity analysis}

Sensitivity analysis (SA) algorithms are an important instrument in the analysis of the models and modeling results. It can be used to estimate how uncertainty in the input parameters affects the output. SA reduces the evaluation cost by removing the least important features from data. This approach is successfully used in different methods of feature importance estimation. 

However, SA also can be used to analyze the structure of composite pipelines obtained by automated methods. Such analysis is important since the pipelines have a heterogeneous graph-based structure containing different operations, which can be redundant or non-optimal. The evolutionary approach described in the Subsec~\ref{subsec_evo} can produce a lot of sub-optimal solutions that required additional post-processing. Also, it is useful to estimate the importance of each block in the pipeline.

The idea is to analyze the impact (positive or negative) of each block of the pipeline for the specified modeling tasks. Since most of the AutoML solutions have linear pipelines, there is little coverage of such an analysis in the literature. 


The numerical results of SA represents the ratio of origin quality score to score with node deletion (or replacement) operations. It can be represented as follows:

\begin{equation} 
\label{eq:sensitivity_index_del}
{S}^{imp}_{i} = \frac{1}{N}(1- {\sum_{n=1}^{N}\frac{F({P}^{G'})}{F({P}^{G})}}),
\end{equation}

where $F$ - objective function for pipeline (e.g. modelling error measure), ${P}^{G'}$ - pipeline with modified structure, $N$ - number of SA iterations, $i$ - index of node to analyze.

The larger value of the ${S}^{imp}_{i}$ index, the more importance can be estimated for the  $i$-th block in the pipeline. If the value of the structural importance is negative, it allows finding a modification of the pipeline that allows improving the overall quality score. It can be used to post-process the pipeline and improve its effectiveness. 

\begin{figure}[h]
    \centering
    \includegraphics[width=0.7\columnwidth]{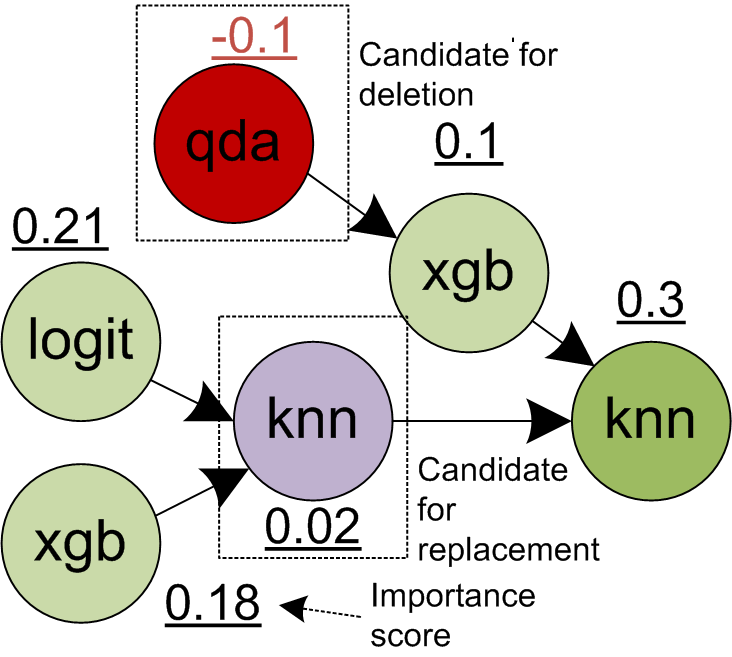}
    \caption{The example of importance analysis for the composite modelling pipeline. The numbers represents the values of ${S}^{imp}$, the colors represent the relative importance of the operations in pipeline.}
    \label{fig_sa}
\end{figure}

This approach allows us to conduct the analysis of the pipeline and obtain the estimations of importance for each node. The example is presented in Fig.~\ref{fig_sa}. 

To confirm the correctness of the proposed SA-based approach for the pipeline analysis, we conducted several simple experiments (the SA was also used during benchmarking in Sec~\ref{sec_exp}). The experiments allow assessing how changes in a single node affect the resulting composite model structure. The experiments were based on the credit scoring (classification)\cite{creditScoring} and cholesterol prediction (regression)\cite{cholesterolRegr} tasks with the static pipeline, static composite pipelines obtained manually and designed by the evolutionary approach described in Subcec.~\ref{subsec_evo}. 

The sustainability index shows how stable the pipeline after structure perturbation, that is, what part of the original composite model remains after its size reduction. The analysis results for the static composite pipeline in the classification problem show that in four of six cases deleting of the component is not recommended. Moreover, the node replacement (with all possible models within the problem) is better, on average, only in one case, out of six. The evolutionary obtained pipeline within 20 generations and populations in size of 20 shows the absolute stability. In other words, no candidates for replacement or deletion are found. The linear pipeline created manually is also stable enough.

The result for the regression problem shows that the pipelines obtained statically have low structural quality because SA results confirm that the removal or replacement of several nodes leads to an increase in the score. On the opposite side, the composed pipeline for the regression problem is much more robust even on a low number of generations. The results for the comparison of different pipelines are presented in Tab.~\ref{tab_sa}.

\begin{table}
\centering
\caption{Results of experiment with sensitivity analysis different implementation of pipelines. Sustainability index is the measure of pipeline stability (the higher - the better), $N_{total}$  - number of nodes in pipeline, $N_{repl}$ - number of candidate nodes for replacement, $N_{del}$ - number of candidate nodes for deletion.}
\label{tab_sa}
\begin{tabular}{|c|c|c|c|c|c|}
\hline
Case                   & Pipeline  & \begin{tabular}[c]{@{}c@{}}Sustainability \\ index\end{tabular}  & $N_{del}$ & $N_{repl}$ & $N_{total}$ \\ \hline
\multirow{3}{*}{Class} & Linear    & 0.8 & 1    & 4     & 5      \\ \cline{2-6} 
                       & Static    & 0.7 & 2    & 1     & 6      \\ \cline{2-6} 
                       & Composite & 1   & 0    & 0     & 5      \\ \hline
\multirow{3}{*}{Reg}   & Linear    & 0.4 & 3    & 1     & 5      \\ \cline{2-6} 
                       & Static    & 0.4 & 3    & 1     & 5      \\ \cline{2-6} 
                       & Composite & 1   & 0    & 0     & 3      \\ \hline
\end{tabular}
\end{table}

The proposed implementation of the sensitivity analysis can contribute to composite pipeline structure explanation and also improve the modeling quality. The described approach is available as a part of the FEDOT framework and described in the tutorial at the framework official documentation\footnote{https://fedot.readthedocs.io/en/latest/fedot/features/sensitivity\_analysis.html}.

\subsection{Reproducilibity of pipelines}

The ability to reproduce the results of experiments in the field of computer science is essential both from theoretical and practical points of view \cite{peng2011reproducible}. The field of ML pipelines requires special attention to the reproducibility of experiments because the fitting of pipelines using large datasets takes a significant amount of time and is highly dependent on the technical capabilities of the system.

There is a set of requirements that can be proposed in terms of composite pipelines reproducibility \cite{tatman2018practical}. It allows defining the set criteria for both data and ML models. The reproducibility from the input data point of view can be achieved by the saving of the sample used for training the models as a set of files. The reproducibility of ML models can be achieved by the preservation of the ML methods and models, the version of the programming language, the versions, and the names of the external libraries. Also, the image of the virtual machine or a whole file system can be created \cite{pineau2020improving}. 

 Most of the state-of-the-art AutoML frameworks have a quite simple implementation of the pipeline exports (as script or binary file with serialized model). For the WMS, the common way for workflow exporting is a description in the domain-specific language. We combine the advantages of both approaches, so the proposed approach includes the specific features to ensure reproducibility. Also, the existing methods for representation of tree-based results of genetic programming \cite{dou2020gpml} were taken into account. To provide a consistent import and export of pipelines in different systems, its structural representation is implemented in human-readable JSON format. The data exporting is implemented as an archive that contains both train and validation data. Also, the interface for creating and managing a virtual machine can be implemented if necessary. It makes it possible to create virtual containers with the necessary dependencies, that allow us to build composite pipelines and fit them on different datasets.

\definecolor{background}{HTML}{FFFFFF}
\lstdefinelanguage{json}{
    basicstyle=\small\ttfamily,
    numberstyle=\scriptsize,
    stepnumber=1,
    numbersep=2pt,
    showstringspaces=false,
    breaklines=true,
    frame=lines,
    backgroundcolor=\color{background},
    moredelim=**[is][\color{red}]{@}{@}
}

\begin{lstlisting}[language=json,mathescape=true,label={lst:fig_json},caption={Ensemble architecture in JSON fromat}]
{
# $\textit{TYPES OF THE NODES IN THE ENSEMBLE}$
"total_pipeline_operations": {
    "xgboost": 1,
    "scaling": 1,
    ...
},
# $\textit{DEPTH OF THE ENSEMBLE}$
"depth": 3,
# $\textit{ALL NODES IN THE ENSEMBLE}$
"nodes": [
    {
        "operation_id": 1,
        "operation_type": "xgboost",
        "operation_name": "XGBClassifier",
        # $\textit{CUSTOM PARAMS}$
        "custom_params": { "n_estimators": 250 },
        # $\textit{FULL PARAMS (custom + default)}$
        "params": {
            "learning_rate": 0.3,
            "max_depth": 6,
            "n_estimators": 250,
             ...
        },
        # $\textit{PARENT NODES IDs}$
        "nodes_from": [0],
        # $\textit{FITTED MODEL PATH}$
        "fitted_operation_path": "fitted_operations/operation_1.pkl"
    },
    ...
]}
\end{lstlisting}

\begin{figure}[h]
    \centering
    \includegraphics[width=\columnwidth]{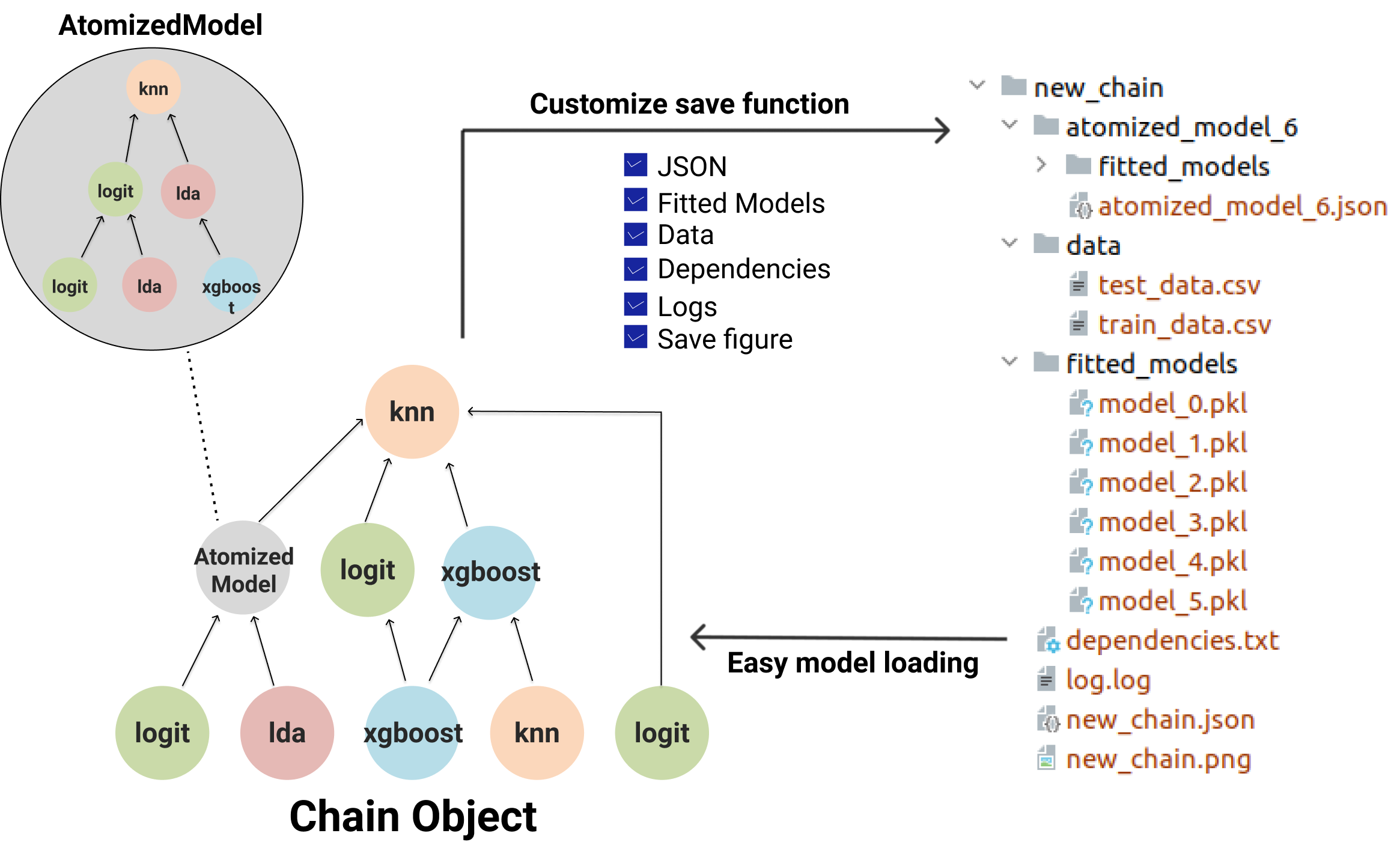}
    \caption{The implemented approach for reproducible of the composite pipelines. The example of atomization for pipeline is also presented.}
    \label{fig_json_atomized}
\end{figure}

The implemented module has a flexible functionality for the communication of the results and their reproducibility in other systems. The user can save the pipeline structure in JSON (Lst.~\ref{lst:fig_json}) format, which describes pipeline structure. In addition, users can save trained models, data, dependencies, logs, and a visual representation of the pipeline as shown in Fig.\ref{fig_json_atomized}. In this case, importing data is as easy as exporting it.

Besides, the pipeline itself can be transformed into an atomized model that can be embedded in the new pipeline as a modeling block, as shown in Fig.\ref{fig_json_atomized}. This allows creating nested pipelines that can be used to achieve higher effectiveness for multi-scale tasks.

\begin{figure*}[h]
    \centering
    \includegraphics[width=0.8\textwidth]{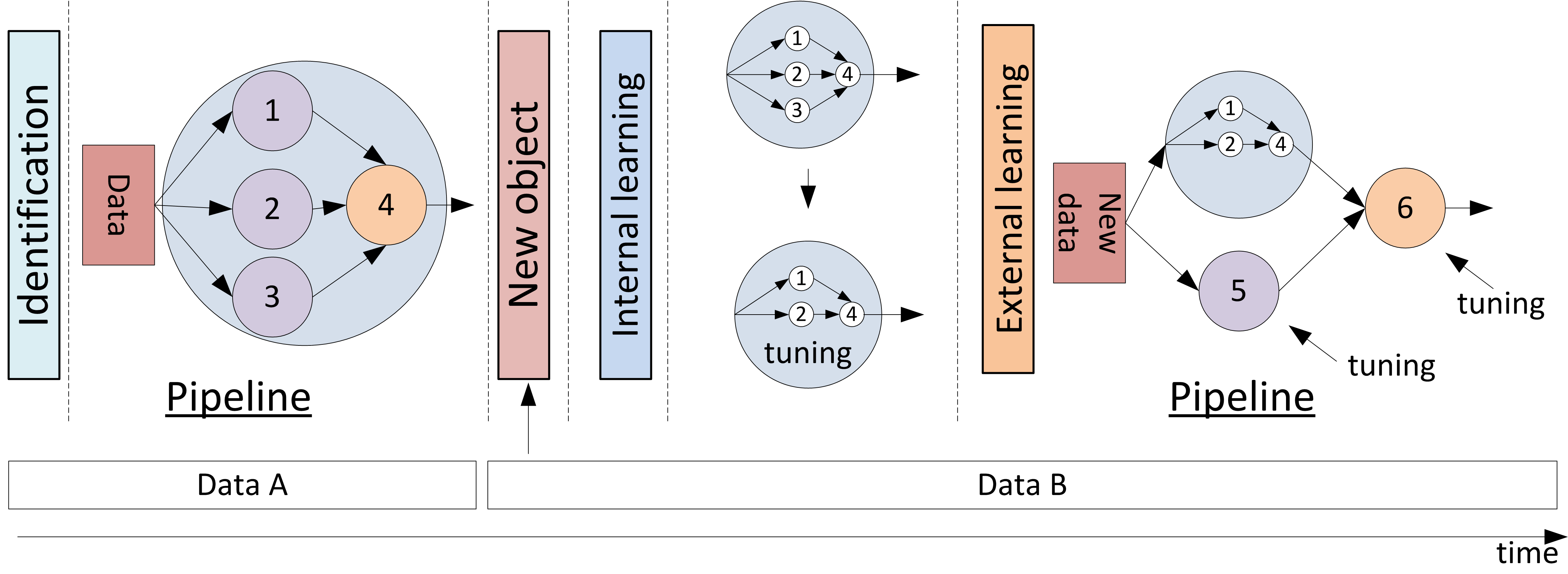}
    \caption{The scheme of the proposed approach for adaptation of the composite pipelines to the new data using atomization. It allow decreasing the search space for the evolutionary optimization and reduce computational requirements.}
    \label{fig_atom}
\end{figure*}

The described atomization technique is also can be used to adapt the pipeline to the updated data sample. In comparison with the pipeline design from the scratch (or from the existing pipeline as an initial assumption), the atomization-based approach allows decreasing the dimensionality of the search space and achieves the appropriate results faster. The idea is described in Fig.~\ref{fig_atom}.

The simple initial experimental evaluation was conducted for the classification task (credit scoring). The initial pipeline with five nodes was identified using evolutionary composer for the first 5000 rows of data  (quality for validation sample is ROC AUC 0.834). Then, 10000 rows of data were added and optimization re-stared from different initial assumptions: previous pipeline and its atomized version). The atomization-based approach allows achieving higher quality (ROC AUC 0.848 vs 0.844) for lower time (3 min vs 8 min). These results were also confirmed for other datasets. For this reason, atomization was used as a part of evolutionary composer and used during experimental studies.

\section{Software implementation}
\label{sec_software}

The proposed approach is implemented as an algorithmic core of the FEDOT framework. It was designed as a multi-purpose AutoML tool that allows us to identify a suitable machine learning pipeline for a given task and dataset in an automatic way. The framework is not focused on certain AutoML subtasks, such as data preprocessing, feature selection, or hyperparameters optimization, but allows one to solve a general task such as structure learning. In this case, for a given dataset the solution is presented by a directed acyclic graph (DAG) structure, where the nodes are ML models or data operations and the edges are dataflows between the nodes.

The architecture of the framework is based on several principles. Firstly, the core logic (especially, the optimization process) should not be strictly dependent on a certain task. Secondly, there is an option to simply extend or replace most of the modules in the framework for the custom user requirements.

In order to simplify the usage of the framework, a high-level Python API was implemented. It is inspired by well-known frameworks like H2O or AutoGluon and makes it possible to obtain an ML pipeline with only several lines of code. The users can customize and instantiate FEDOT objects for their needs.     

\subsection{ML pipelines execution}

In FEDOT, ML pipeline is considered as a \textsl{Pipeline} object. It is an abstraction of a DAG structure that provides interfaces for graph modifications and pipeline execution. Pipeline contains \textsl{Nodes} that isolate implementations of certain operations. The execution of the pipeline is initialized recursively from a final node, and the data flow is going in the opposite direction.

Each node includes \textsl{Operation} object that is a base class of two interfaces - \textsl{Model} and \textsl{DataOperation}. In ML pipelines a particular model or data operation can be considered as building blocks. In general, model is a function that transform input data (features) into output (predictions) and has two methods - \textsl{fit} and \textsl{predict}. In turn, data operations modify only input data, like PCA, undersampling, or normalization. 

Also, the same operations can be implemented in various ways. Therefore we additionally designed a \textsl{EvaluationStrategy} logic layer in the framework. For instance, if the user wants to use a logistic regression from scikit-learn\footnote{https://github.com/scikit-learn/scikit-learn} framework, then it must be implemented as a \textsl{SklearnEvaluationStrategy} with logistic regression inside in FEDOT.

\subsection{ML pipelines building and optimization}

Pipeline structure optimization logic is embedded into a separate module. In FEDOT the entry point for this operation is a \textsl{Composer} interface. Thus, the optimization is called \textsl{composing}. The purpose of a Composer is to connect pipeline-related logic and optimization algorithms. The last are implemented via sub-classes of \textsl{Optimiser} interface. For instance, previously mentioned GPComp algorithm is provided as a \textsl{GPOptimiser} class. The described design allows us to conduct the experiments with optimization algorithms of different nature without strong connectivity with the core of FEDOT.

\begin{figure*}[t]
\centerline{\includegraphics[width=15cm]{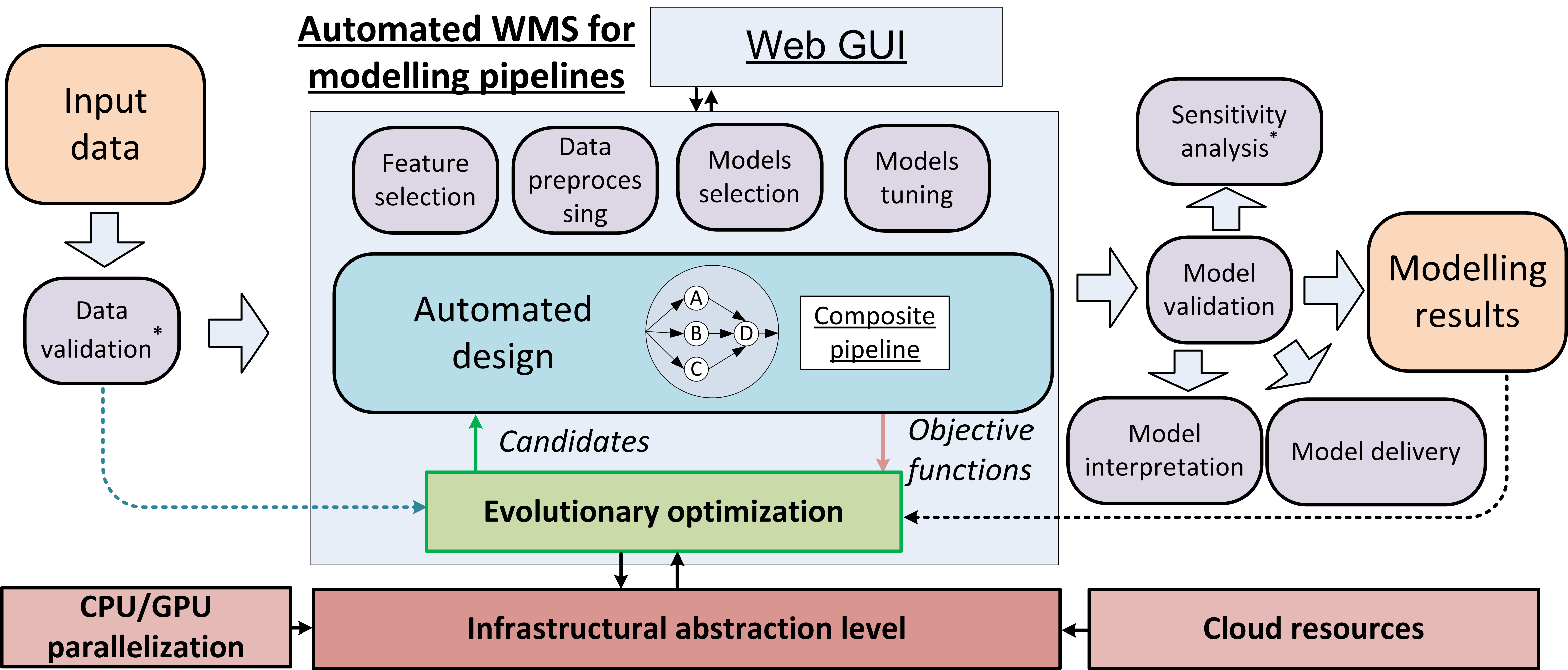}}
\caption{The structure of proposed solution for automated design of the modelling workflows. The implementation of this approach is available as a core of FEDOT framework}
\label{fig_detailed_pipeline}
\end{figure*}

\section{Experimental studies for efficiency analysis of the proposed approached}
\label{sec_exp}


\subsection{Classification and regression benchmarks}

To demonstrate the effectiveness of the evolutionary design of pipelines, an experiment was conducted using ten data sets for regression and classification problems (five data sets for each problem), obtained from the Penn Machine Learning Benchmarks repository\footnote{https://github.com/EpistasisLab/pmlb}. These data sets cover a broad range of applications, and combinations of categorical, ordinal, and continuous features. There are no missing values in these data sets. Selected data sets were split on training and test set in a ratio of 80/20. An example of running an experiment is located in the specified repository.\footnote{https://github.com/ITMO-NSS-team/FEDOT-benchmarks/blob/master/experiments/four\_pipelines/run.py}

The values of quality metrics are presented in Table~\ref{tab_automl_comprasion}. The experiment includes:
\begin{itemize}
\item Initialisation of several tools: FEDOT framework implementation (based on the proposed approach), TPOT and MLBox frameworks, baseline model (XGBoost);
\item Evaluation of the automated pipeline design using each approach on the set of classification and regression benchmarks;
\item Estimation of modeling error: MAE and RMSE are used as quality metrics for regression tasks and F1, ROC-AUC is used as a quality metric for classification tasks.
\end{itemize}

To initialize the setups for evolutionary tools (TPOT and FEDOT), the values of the following hyperparameters were selected: a maximum number of generations to create a composite pipeline – 200; population size – 10 individuals. The maximum time for evaluation was limited by 10 min for all approaches.

The following hyperparameters values were selected for the baseline XGBoost model: maximum depth - 3; shrinkage coefficient of each tree contribution – 0.3; the number of boosting stages to perform – 300.

The main meta-features of the selected data sets are shown in Table \ref{tab_dataset_features}. Imbalance shows a value of imbalance metric, where zero means that the data set is perfectly balanced, and the higher the value, the more imbalanced the data set.

\begin{table}
\centering
\caption{Main meta-features of the PMLB datasets for regression (regr) and classification (clf) used during experimenents.}
\label{tab_dataset_features}
\begin{tabular}{|c|c|c|c|c|} 
\hline
Dataset   & $N_{samples}$ & $N_{feat}$ & Task & Imbalance  \\ 
\hline
Cpu\_small & 8192          & 12             & regr.       & -          \\ 
\hline
Elusage    & 55            & 2              & regr.       & -          \\ 
\hline
Faculty    & 50            & 4              & regr.       & -          \\ 
\hline
C2\_250\_25  & 250           & 25             & regr.       & -          \\ 
\hline
1027\_ESL  & 488           & 4              & regr.       & -          \\ 
\hline
Magic      & 19020         & 10             & binary clf. & 0.08       \\ 
\hline
Labor      & 57            & 16             & binary clf. & 0.08       \\ 
\hline
Flare      & 1066          & 10             & binary clf. & 0.43       \\ 
\hline
Ionosphere & 351           & 34             & binary clf. & 0.08       \\ 
\hline
Spect      & 267           & 22             & binary clf. & 0.34       \\
\hline
\end{tabular}
\end{table}

As we can see from Table~\ref{tab_automl_comprasion}, the results obtained during the experiments demonstrate the advantage of composite pipeline created by the proposed approach over competitors. The only exception is a single case for regression and classification problems respectively, where the maximum value of the quality metric was obtained using a static pipeline. However, it should be noted that all 10 experiments with the proposed approach showed better results in comparison with the other AutoML frameworks.

\begin{table*}[ht!]
\centering
\caption{Results of experiment and values of quality metrics for each of the proposed benchmark approaches. The standard deviation of the quality metrics (MAE, RMSE, F1, area under ROC curve) is estimated for the 20 independent runs.}
\label{tab_automl_comprasion}
\begin{tabular}{|c|c|c|c|c|c|c|} 
\hline
\multirow{2}{*}{\begin{tabular}[c]{@{}c@{}}Quality\\metric\end{tabular}} & \multirow{2}{*}{\begin{tabular}[c]{@{}c@{}}Pipeline and \\framework\end{tabular}} & \multicolumn{5}{c|}{\textbf{\textbf{\textbf{\textbf{Regression data sets}}}}}                                                                                                                                                                            \\ 
\cline{3-7}
                                                                         &                                                                                   & 1027\_ESL                                       & 1096\_FacultySalaries                     & 227\_cpu\_small                                             & 228\_elusage                                    & 605\_fri\_c2\_250\_25                      \\ 
\hline
\multirow{4}{*}{MAE}                                                     & Static (XGBoost)                                                                  & 0.441\textcolor[rgb]{0.2,0.2,0.2}{±0.010}       & 1.597\textcolor[rgb]{0.2,0.2,0.2}{±0.027} & 2.006\textcolor[rgb]{0.2,0.2,0.2}{±0.023}                   & 10.484\textcolor[rgb]{0.2,0.2,0.2}{±0.037}      & \textbf{0.349\textbf{±0.019}}              \\ 
\cline{2-7}
                                                                         & \textbf{\textbf{Composite (FEDOT)}}                                               & \textbf{0.430\textbf{±0.009}}                   & \textbf{1.120\textbf{±0.012}}             & \textbf{1.931}\textcolor[rgb]{0.2,0.2,0.2}{\textbf{±0.007}} & \textbf{9.451\textbf{±0.020}}                   & 0.371\textcolor[rgb]{0.2,0.2,0.2}{±0.033}  \\ 
\cline{2-7}
                                                                         & Variable (TPOT)                                                                   & 0.432\textcolor[rgb]{0.2,0.2,0.2}{±0.004}       & 1.508\textcolor[rgb]{0.2,0.2,0.2}{±0.023} & 2.064\textcolor[rgb]{0.2,0.2,0.2}{±0.011}                   & 11.051\textcolor[rgb]{0.2,0.2,0.2}{±0.031}      & 0.373\textcolor[rgb]{0.2,0.2,0.2}{±0.014}  \\ 
\cline{2-7}
                                                                         & Linear (MLBox)                                                                    & 0.434\textcolor[rgb]{0.2,0.2,0.2}{±0.018}       & 3.670\textcolor[rgb]{0.2,0.2,0.2}{±0.048} & 1.934\textcolor[rgb]{0.2,0.2,0.2}{±0.005}                   & 21.075\textcolor[rgb]{0.2,0.2,0.2}{±0.054}      & 0.423\textcolor[rgb]{0.2,0.2,0.2}{±0.076}  \\ 
\hline
\multirow{4}{*}{RMSE}                                                    & Static (XGBoost)                                                                  & 0.588\textcolor[rgb]{0.2,0.2,0.2}{±0.028}       & 2.199\textcolor[rgb]{0.2,0.2,0.2}{±0.037} & 2.799\textcolor[rgb]{0.2,0.2,0.2}{±0.023}                   & 14.053\textcolor[rgb]{0.2,0.2,0.2}{±0.023}      & \textbf{0.448\textbf{±0.025}}              \\ 
\cline{2-7}
                                                                         & \textbf{Composite (FEDOT)}                                                        & \textbf{0.566\textbf{±0.013}}                   & \textbf{1.403\textbf{±0.024}}             & \textbf{2.759\textbf{±0.007}}                               & \textbf{11.558\textbf{±0.020}}                  & 0.466\textcolor[rgb]{0.2,0.2,0.2}{±0.035}  \\ 
\cline{2-7}
                                                                         & Variable (TPOT)                                                                   & 0.570\textcolor[rgb]{0.2,0.2,0.2}{±0.009}       & 1.789\textcolor[rgb]{0.2,0.2,0.2}{±0.033} & 2.886\textcolor[rgb]{0.2,0.2,0.2}{±0.023}                   & 14.863\textcolor[rgb]{0.2,0.2,0.2}{±0.032}      & 0.465\textcolor[rgb]{0.2,0.2,0.2}{±0.014}  \\ 
\cline{2-7}
                                                                         & Linear (MLBox)                                                                    & 0.628\textcolor[rgb]{0.2,0.2,0.2}{±0.052}       & 4.894\textcolor[rgb]{0.2,0.2,0.2}{±0.061} & 2.793\textcolor[rgb]{0.2,0.2,0.2}{±0.023}                   & 26.141\textcolor[rgb]{0.2,0.2,0.2}{±2.27}       & 0.505\textcolor[rgb]{0.2,0.2,0.2}{±0.089}  \\ 
\hline
\multirow{2}{*}{\begin{tabular}[c]{@{}c@{}}Quality\\metric\end{tabular}} & \multirow{2}{*}{\begin{tabular}[c]{@{}c@{}}Pipeline and\\framework\end{tabular}}  & \multicolumn{5}{c|}{\textbf{\textbf{Classification data sets}}}                                                                                                                                                                                          \\ 
\cline{3-7}
                                                                         &                                                                                   & flare                                           & ionosphere                                & labor                                                       & magic                                           & spect                                      \\ 
\hline
\multirow{4}{*}{F1}                                                      & Static (XGBoost)                                                                  & 0.299\textcolor[rgb]{0.2,0.2,0.2}{±0.110}       & \textbf{0.939\textbf{±0.026}}             & 0.895\textcolor[rgb]{0.2,0.2,0.2}{±0.023}                   & 0.793\textcolor[rgb]{0.2,0.2,0.2}{±0.013}       & 0.837\textcolor[rgb]{0.2,0.2,0.2}{±0.077}  \\ 
\cline{2-7}
                                                                         & \textbf{\textbf{Composite (FEDOT)}}                                               & \textbf{0.332\textbf{\textbf{\textbf{±0.057}}}} & 0.919\textcolor[rgb]{0.2,0.2,0.2}{±0.026} & \textbf{0.931\textbf{±0.013}}                               & \textbf{0.817\textbf{\textbf{\textbf{±0.004}}}} & \textbf{0.893\textbf{±0.062}}              \\ 
\cline{2-7}
                                                                         & Variable (TPOT)                                                                   & 0.292\textcolor[rgb]{0.2,0.2,0.2}{±0.034}       & 0.774\textcolor[rgb]{0.2,0.2,0.2}{±0.052} & 0.840\textcolor[rgb]{0.2,0.2,0.2}{±0.023}                   & 0.809\textcolor[rgb]{0.2,0.2,0.2}{±0.004}       & 0.830\textcolor[rgb]{0.2,0.2,0.2}{±0.046}  \\ 
\cline{2-7}
                                                                         & Linear (MLBox)                                                                    & 0.181\textcolor[rgb]{0.2,0.2,0.2}{±0.097}       & 0.933\textcolor[rgb]{0.2,0.2,0.2}{±0.088} & 0.840\textcolor[rgb]{0.2,0.2,0.2}{±0.069}                   & 0.812\textcolor[rgb]{0.2,0.2,0.2}{±0.017}       & 0.845\textcolor[rgb]{0.2,0.2,0.2}{±0.071}  \\ 
\hline
\multirow{4}{*}{\begin{tabular}[c]{@{}c@{}}ROC\\AUC\end{tabular}}        & Static (XGBoost)                                                                  & 0.693\textcolor[rgb]{0.2,0.2,0.2}{±0.025}       & \textbf{0.956\textbf{±0.007}}             & 0.923\textcolor[rgb]{0.2,0.2,0.2}{±0.023}                   & 0.915\textcolor[rgb]{0.2,0.2,0.2}{±0.015}       & 0.736\textcolor[rgb]{0.2,0.2,0.2}{±0.075}  \\ 
\cline{2-7}
                                                                         & \textbf{\textbf{\textbf{\textbf{Composite (FEDOT)}}}}                             & \textbf{0.708\textbf{±0.033}}                   & 0.951\textcolor[rgb]{0.2,0.2,0.2}{±0.011} & \textbf{0.958\textbf{±0.019}}                               & \textbf{0.930\textbf{±0.004}}                   & \textbf{0.779\textbf{±0.043}}              \\ 
\cline{2-7}
                                                                         & Variable (TPOT)                                                                   & 0.701\textcolor[rgb]{0.2,0.2,0.2}{±0.008}       & 0.954\textcolor[rgb]{0.2,0.2,0.2}{±0.079} & 0.958\textcolor[rgb]{0.2,0.2,0.2}{±0.023}                   & 0.928\textcolor[rgb]{0.2,0.2,0.2}{±0.004}       & 0.657\textcolor[rgb]{0.2,0.2,0.2}{±0.039}  \\ 
\cline{2-7}
                                                                         & Linear (MLBox)                                                                  & 0.509\textcolor[rgb]{0.2,0.2,0.2}{±0.012}       & 0.907\textcolor[rgb]{0.2,0.2,0.2}{±1.036} & 0.515\textcolor[rgb]{0.2,0.2,0.2}{±0.058}                   & 0.856\textcolor[rgb]{0.2,0.2,0.2}{±0.026}       & 0.628\textcolor[rgb]{0.2,0.2,0.2}{±0.031}  \\
\hline
\end{tabular}
\end{table*}

The obtained results confirm the correctness, effectiveness, flexibility, and competitiveness of the proposed approach and its implementation in the FEDOT framework.

\subsection{Time series forecasting benchmarks}

An important task in machine learning is time series forecasting, but just a few libraries for automatic machine learning can successfully perform it. The proposed approach can be also used for the automated design of the pipelines for the time series forecasting.

During the experiments, a comparison with specialized libraries for automated time series forecasting (AutoTS~\cite{khider2019autots} and Prophet~\cite{taylor2018forecasting}) was conducted.

For comparison, twelve time-series datasets obtained from the archive of Federal Reserve Economic Data (FRED) were used. Ten datasets had a length from 408 to 1028 elements - they are considered as examples of short series during experiments. Two of them had a length of 17000 elements - they are considered as examples of long series. The forecast horizon was changed from 10 to 100 elements with a step of ten elements for short series, and from 10 to 200 elements with the same step for long time series. For each forecast, the predicted and actual values were compared using the Mean Absolute Percentage Error (MAPE). The execution time of each library for finding a solution was also monitored and limited. In some cases, algorithms were run until convergence was achieved.

The experimental results allow confirming that the proposed approach outperforms competitors (Table~\ref{ts_errors}) in most cases.

\begin{table}[ht!]
\caption{Results of an experiment with the time series forecasting The modeling error is represented as MAPE. $h$ is the forecast horizon. The quality metric (MAPE) is averaged for the 20 independent runs. The composite pipeline is obtained automatically using the proposed approach.}
\label{ts_errors}
\centering
\begin{tabular}{|c|c|c|c|c|c|} 
\hline
\multirow{3}{*}{Tool} & \multicolumn{5}{c|}{\textbf{\textbf{MAPE for validation part, \%}}} \\ 
\cline{2-6}
 & \multicolumn{2}{c|}{Short time series} & \multicolumn{2}{c|}{Long time series} & \multirow{2}{*}{Avg.}  \\ 
\cline{2-5}
& h $\leq$ 50 & h $\leq$ 100 & h $\leq$ 100 & h $\leq$ 200 & \\ 
\hline
Prophet & 14.6 & 12.7 & 78.7 & 95.6  & 50.4 \\ 
\hline
AutoTS  & 7.1 & 11.7 & 25.9 & 18.7 & 15.8 \\ 
\hline
\begin{tabular}[c]{@{}@{}c@{}}\textbf{Comp.}\\\textbf{pipeline}
\\\textbf{(FEDOT)}\end{tabular} & \textbf{6.8} & \textbf{10.7}           & \textbf{9.6} & \textbf{6.6} & \textbf{8.4} \\
\hline
\end{tabular}
\end{table}

\begin{figure}[h]
    \centering
    \includegraphics[width=1.0\columnwidth]{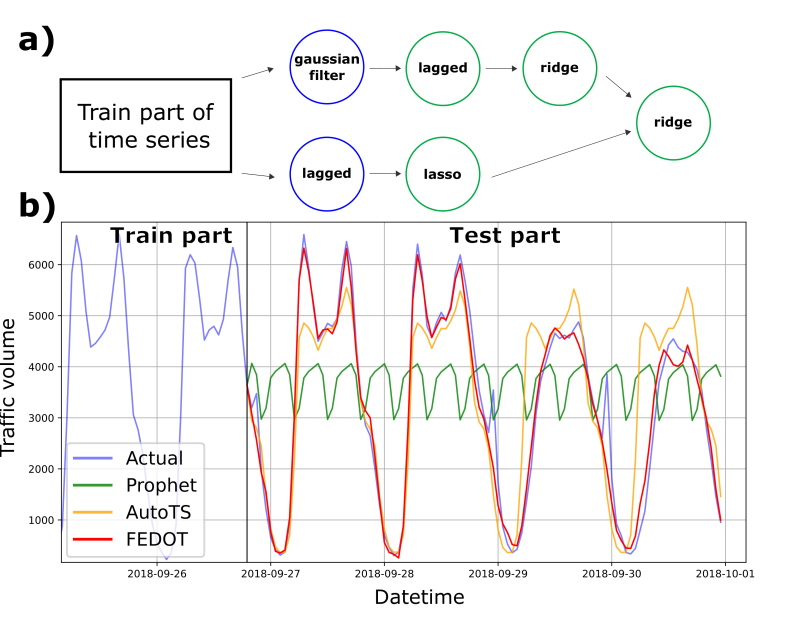}
    \caption{The details of the composed pipeline for time series forecasting task a) Structure of the pipeline; b) Comparison between forecast obtained with evolutionary obtained pipeline and competing libraries forecasts (for test part of time series)}
    \label{ts_pipeline}
\end{figure}

As can be seen from Table~\ref{ts_errors}, the proposed approach demonstrates the significant advantage against state-of-the-art libraries for all forecasting horizons and time series. Fig.~\ref{ts_pipeline} demonstrates an example of the composing pipeline obtained for the time series forecasting task. As can be seen from the Fig.~\ref{ts_pipeline}, the pipeline combines several preprocessing blocks and modeling blocks.

The scripts and data for described experiments with time series forecasting are available in the repository \footnote{\url{https://github.com/ITMO-NSS-team/Fedot-TS-Benchmark/tree/master/time_series_case_1}}.

\subsection{Analysis of computational effectiveness}

The previous sub-sections were devoted to the analysis of the effectiveness of the proposed approach. At the same time, the analysis of computational properties of the underlying algorithm is also an important part of the experimental studies. The execution time limit for the experiments was fixed (to achieve a fair comparison with other solutions), so the memory consumption and convergence of the implemented algorithms are analyzed.

\begin{figure}[h]
    \centering
    \includegraphics[width=1.0\columnwidth]{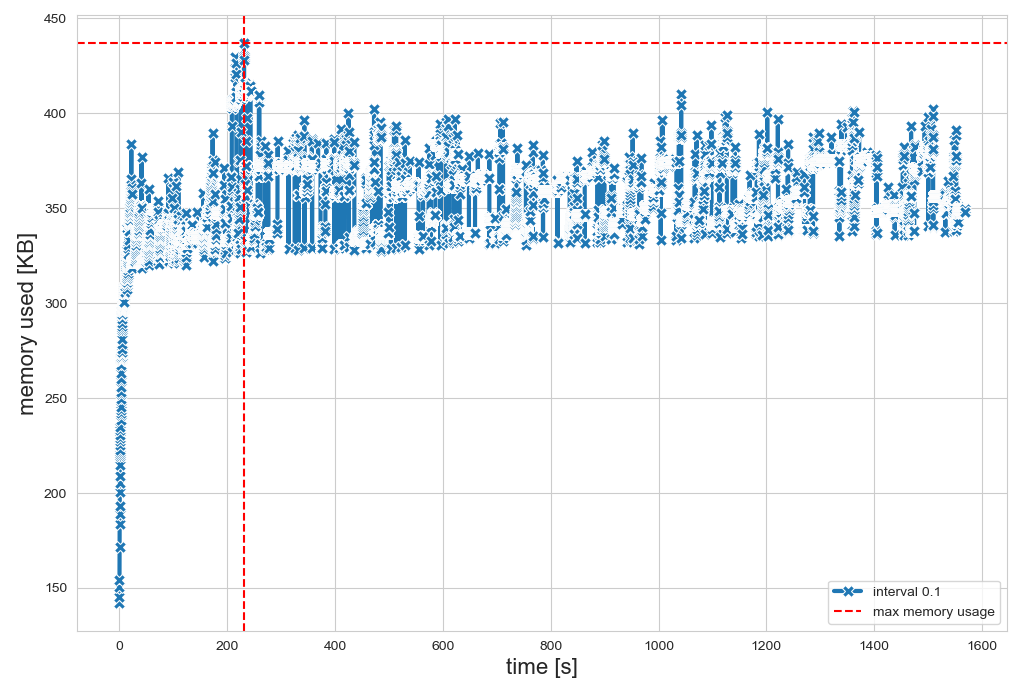}
    \caption{The changes of system memory consumption for the model design process using the proposed approach. The length of the optimization run was restricted with 30 min limit. The red line represents the value of maximum memory usage, observed during the experiment.}
    \label{fig_memory}
\end{figure}

The memory consumption analysis was conducted using the custom tool (implemented as a part of FEDOT) that is based at 'memory\_profiler' library \footnote{\url{https://github.com/pythonprofilers/memory_profiler}}. The set of benchmarks were used to confirm that value of used memory is stable during the optimization run (the memory-related problems can be caused by memory leaks, excessive copy operations, etc). The example of analysis results is presented in Fig.~\ref{fig_memory}.

It can be seen that sharp changes in memory consumption occur, which can be explained by the pipelines fitting for each individual in the population during the fitness evaluation. The long-term memory usage is stable, which can confirm the correctness and effectiveness of the algorithmic implementation of the proposed approach.

\begin{figure}[t!]
\centerline{\includegraphics[width=0.45\textwidth]{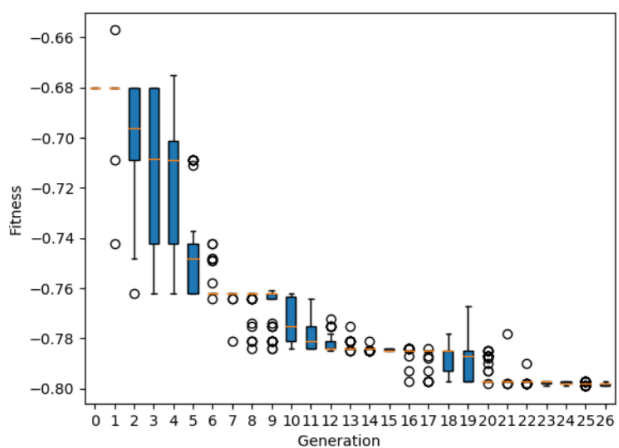}}
\caption{The boxplots for fitness function in different generations for the evolutionary design started from initial assumption (the dataset 'Magic' is used).}
\label{fig_fitnes_eval}
\end{figure}

The convergence of the implemented evolutionary algorithms also was analyzed. Fig.~\ref{fig_fitnes_eval} presents the dynamic of the fitness functions value for optimization run started from the initial assumption.

As seen from Fig.~\ref{fig_fitnes_eval}, with increasing the number of generations, the variance of the fitness function value decreases, and the average fitness improves, which indicates that the optimization algorithm is converging successfully.

So, we can conclude that the proposed approach is implemented in a computationally efficient way and has no significant computational overheads. The demonstrated convergence of the evolutionary optimization confirms the stability of the results of the automated pipeline identification. The results from previous sections allow claiming that the proposed approach can achieve better results with the same time and computational resources, which makes it a promising candidate for the processing of complicated machine learning pipelines.

\section{Discussions and conclusions}
\label{sec_disc}


In the paper, the evolutionary approach for the automated design of the data-driven modeling workflow is proposed. It is based on the concept of a combination of pipeline design methods used AutoML and workflow management systems: graph-based structure of pipeline, several types of building blocks, descriptive representation of the pipeline, specialized methods for fine-tuning, structural explainability, etc.

This approach allows building graph-based pipelines that consist of different building blocks: data-driven models, data prepossessing functions, task-specific models, data flow transformations. Each pipeline is designed individually for a specific task that allows reducing the time efforts and increases the modeling quality. It is effective for the modeling tasks with high uncertainty, multi-scale character, or in the cases with raw unprocessed data available only. However, for the simple benchmarks, it can be redundant since a larger time is required to evaluate a larger search space for pipeline design.

The implemented library can be integrated into existing WMS for machine learning in different ways: the simplest one is integration as an atomic block of pipelines that allow building the models for different tasks; the more advanced is the usage as a part of automation block for the design of pipelines. Also, it can be used as a tool for the co-design of the pipeline and infrastructure if the objective functions for automated search depend on the performance and scalability of the obtained solution.

The proposed approach can be extended to improve the effectiveness of the automation in different computationally-intensive modeling systems that are used in modern e-Science. The proposed approach can be extended to improve the effectiveness of the automation in different computationally-intensive modeling systems that are used in modern e-Science. Additional building blocks, design patterns, optimization heuristics, supported tasks, and data types. However, the common idea described in the paper remains the same.

Besides the algorithmic development, we make this approach available as a part of the self-developed AutoML framework - FEDOT. It is an open-source tool that can be used under a free BSD-3 clause license. There is both a simple interface (API) for inexperienced users and a customizable API for experts are available. The evolutionary optimizer is used to automate the pipeline design. To achieve better effectiveness and applicability of the proposed approach, additional procedures for sensitivity analysis and atomization are implemented.

The experimental validation of the proposed approach was conducted on different tasks. The classification, regression, and time series forecasting benchmarks were evaluated for the set of state-of-the-art solutions and baselines. The FEDOT-based implementation of the evolutionary approach demonstrates a significant advantage over competitors.

We evaluate our approach only for a limited set of benchmarks and do not claim to have the superior approach for all tasks. In practice, there is a wide variety of effective AutoML solutions exists. Some of them have flaws in technical aspects (such as support for scalability and distributed computing, Kubernetes, and integration with MLOps tools). In others, there are conceptual issues like oversimplified optimization algorithms or the lack of their interpretability. So, existing solutions have a space to improve.

For example, there are several areas and prospects of development of AutoML that a based on the concept of 'model factory' that can provide different solutions to the user depending on the given conditions: the types of data sets, forecasting horizons, the lifetime of the model, etc. The modeling pipelines hat models can be derived for different data sets: random samples, data within time ranges. It is also possible to obtain "short-lived" models on a current data slice. We aimed the described approach as a promising step to the applied implementation of this concept. The generative evolutionary design of the pipelines for the different tasks makes it possible to obtain a valuable contribution to the different fields.

\section{Code and data availability}
\label{sec_code}

The software implementation of all described methods and algorithms are available in the open repository \url{https://github.com/nccr-itmo/FEDOT}. The reproducible setups for all presented experiments are available in the \url{https://github.com/ITMO-NSS-team/FEDOT-benchmarks}.

\section{Acknowledgments}

This research is financially supported by the Ministry of Science and Higher Education, Agreement \#075-15-2020-808.

\bibliography{sample}

\end{document}